\documentclass{article}

\usepackage{microtype}
\usepackage{graphicx}
\usepackage{subcaption}
\usepackage{booktabs} % for professional tables

\usepackage{hyperref}

\usepackage[accepted]{icml2026}

\usepackage{amsmath}
\usepackage{amssymb}
\usepackage{mathtools}
\usepackage{amsthm}

\usepackage[capitalize,noabbrev]{cleveref}

\theoremstyle{plain}

\theoremstyle{definition}

\theoremstyle{remark}

\usepackage[textsize=tiny]{todonotes}

\definecolor{lightyellow}{rgb}{1, 1, 0.918}
\definecolor{lightblue}{rgb}{0.918, 0.961, 1}
\definecolor{light}{rgb}{0.918, 0.961, 1}
\usepackage{adjustbox} % for \adjustbox command
\usepackage{multirow}
\usepackage{colortbl}
\usepackage{float}
\usepackage{tocloft}
\usepackage{fontawesome}
\usepackage{bbding}
\usepackage{url}
\usepackage{algorithm}

\newcommand{\vc}{{\bf c}}

\icmltitlerunning{%Interpretable Open-World Object Detection: Redefining the Boundary Between Known and Unknown
Knowing the Unknown: Interpretable Open-World Object Detection via Concept Decomposition Model
}

\begin{document}

\twocolumn[
  \icmltitle{
  % Interpretable Open-World Object Detection: Redefining the Boundary Between Known and Unknown
  Knowing the Unknown: Interpretable Open-World Object Detection via Concept Decomposition Model
  }



  \icmlsetsymbol{equal}{*}

  \begin{icmlauthorlist}
    \icmlauthor{Xueqiang Lv}{nwpu}
    \icmlauthor{Shizhou Zhang}{nwpu}
    \icmlauthor{Yinghui Xing}{nwpu}
    \icmlauthor{Di Xu}{huawei}
    \icmlauthor{Peng Wang}{nwpu}
    \icmlauthor{Yanning Zhang}{nwpu}
    % \icmlauthor{Firstname2 Lastname2}{equal,yyy,comp}
    % \icmlauthor{Firstname3 Lastname3}{comp}
    % \icmlauthor{}{sch}
    % \icmlauthor{}{sch}
    % \icmlauthor{}{sch}
  \end{icmlauthorlist}


  \icmlaffiliation{nwpu}{School of Computer Science, Northwestern Polytechnical Unviersity, Xi'an, China}
  \icmlaffiliation{huawei}{Huawei Technologies Ltd}

\icmlcorrespondingauthor{Shizhou Zhang}{szzhang@nwpu.edu.cn}
\icmlcorrespondingauthor{Yinghui Xing}{xyh\_7491@nwpu.edu.cn}

  \icmlkeywords{Machine Learning, ICML}

  \vskip 0.3in
]



\printAffiliationsAndNotice{}  % no special notice (required even if empty)

\begin{abstract}
% Setting
Open-world object detection (OWOD) requires incrementally detecting known categories while reliably identifying unknown objects.
% 现存方法的缺点
Existing methods primarily focus on improving unknown recall, yet overlook interpretability, often leading to known--unknown confusion and reduced prediction reliability.
% 目标
This paper aims to make the entire OWOD framework interpretable, enabling the detector to truly ``knowing the unknown.''
% 方法
To this end, we propose a concept-driven \textbf{I}nter\textbf{P}retable \textbf{OW}OD framework(IPOW) by introducing a Concept Decomposition Model (CDM) for OWOD, which explicitly decomposes the coupled RoI features in Faster R-CNN into discriminative, shared, and background concepts. 
Discriminative concepts identify the most discriminative features to enlarge the distances between known categories, while shared and background concepts, due to their strong generalization ability, can be readily transferred to detect unknown categories.
Leveraging the interpretable framework, we identify that known--unknown confusion arises when unknown objects fall into the discriminative space of known classes. 
To address this, we propose Concept-Guided Rectification (CGR) to further resolve such confusion.
% 结果
Extensive experiments show that IPOW significantly improves unknown recall while mitigating confusion, and provides concept-level interpretability for both known and unknown predictions.

\end{abstract}
    
\section{Introduction}
% 概述一下open world的setting
Traditional object detection approaches are typically built on a closed-set assumption, limiting detection to known categories that have been seen during training. Open-World Object Detection (OWOD) first introduced by~\cite{joseph2021towards}, overcomes this constraint by defining a new framework that can automatically identify previously unseen objects and gradually learn to classify them through subsequent incremental learning stages, thereby enabling continuous adaptation in real-world settings.
\begin{figure}
	\centering
	\includegraphics[width=1\linewidth]{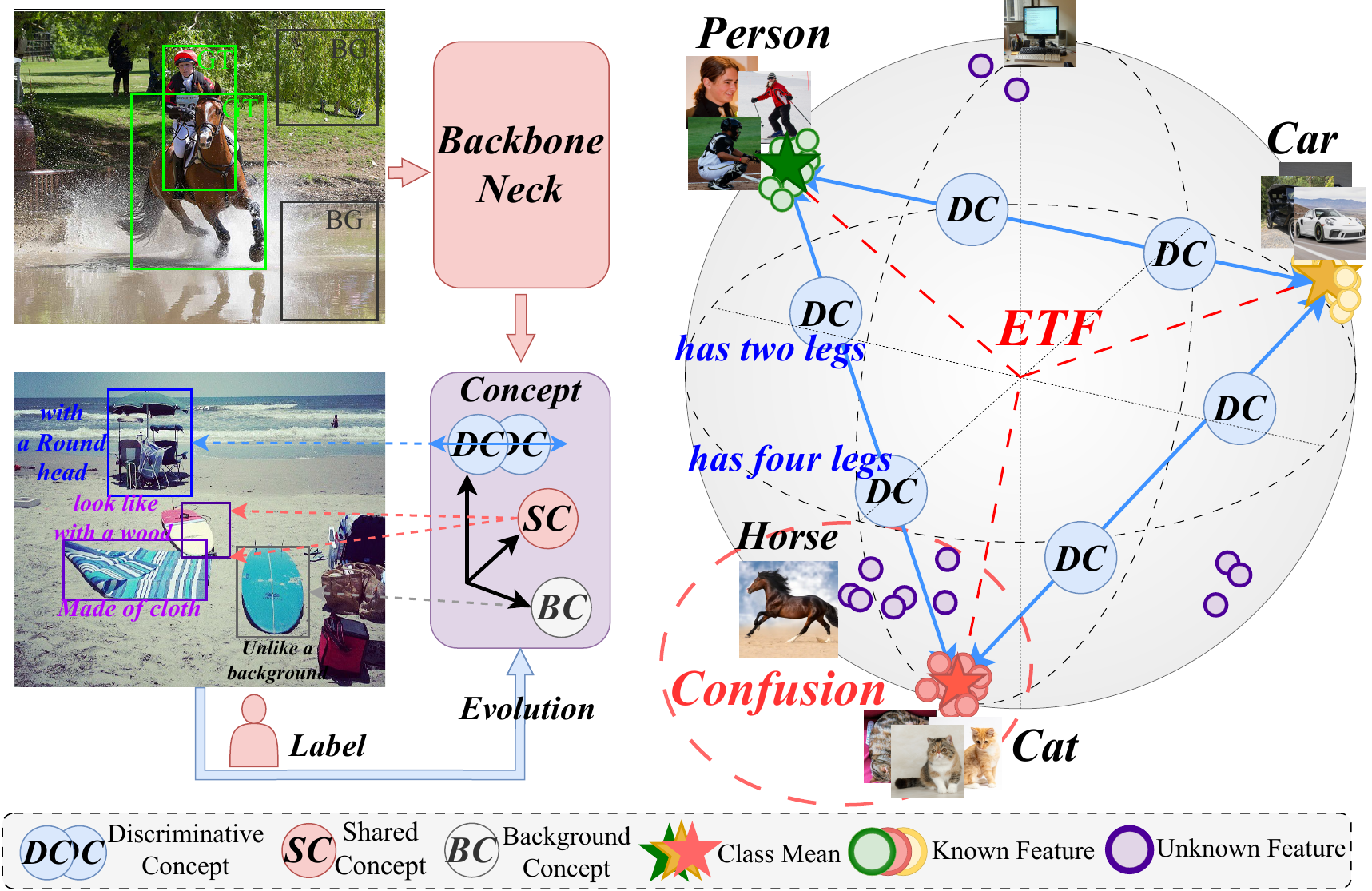}
\caption{\textbf{Left:} The proposed IPOW reformulates OWOD as a problem of concept decomposition modeling, consisting of discriminative concepts responsible for known-class recognition, along with shared and background concepts that support unknown object detection. \textbf{Right:} Within the IPOW framework, known–unknown confusion can be viewed as unknown objects falling into the discriminative space learned for known classes.}
\label{fig:motivation}
    \vspace{-3mm}
\end{figure}
% 概述一下提出OWOD的两个挑战, 已知与未知的混淆 与 已知类别偏差.
However, this shift in paradigm brings about two core challenges: \emph{known--unknown confusion}, where visually similar unknown objects are mistakenly detected as known categories, causing a high rate of false positives; and \emph{bias toward known categories}, where detectors trained exclusively on known classes naturally prioritize these known objects, thereby restricting generalization to unknown classes and yielding low recall on unknown objects.

% 现有方法存在的问题 一是没有探究混淆的本质  二 是依赖于挖掘导致未知recall不够高
Although there have been recent attempts to tackle these issues, substantial difficulties still persist. 
% We attribute the aforementioned problems to insufficient transfer of knowledge from known categories to previously unseen ones.
Specifically, existing methods~\cite{gupta2022ow, majee2025looking} typically employ a class-agnostic objectness head to score all RoIs, heuristically treating high-objectness regions outside known classes as unknown, and predominantly rely on \textit{self-supervised mining based on objectness} to identify latent unknown objects in the training set.
Note that the unknown classes basically consist of two sources. One is \textit{Known Unknown Classes (KUCs)}~\cite{geng2020recent} which denote the unlabeled categories in the training set that cover only a small fraction of the open world. The other is the far more numerous \textit{Unknown Unknown Classes (UUCs)}, which never appear during training, are more common and critical in realistic open-world settings.
As self-supervised mining is limited to KUCs, existing methods often misinterpret background regions as latent unknowns, leading to excessive false positives and low unknown recall.
Additionally, they represent objectness only in an abstract way and their decision-making procedures do not clarify why certain regions are designated as unknown. More critically, they do not account for the reasons behind confusion between known and unknown classes, which undermines the reliability of their predictions.

In this paper, we employ Concept Bottleneck Models to achieve interpretable knowledge transfer from known to unknown classes, directly addressing the above challenges.
% In this paper, we resort to Concept Bottleneck Models to make the knowledge transfer from known to unknown interpretable, thereby fundamentally addressing the challenges outlined above. 
To this end, as shown in Fig.~\ref{fig:motivation}, we propose a concept-driven \textbf{I}nter\textbf{P}retable \textbf{O}pen-\textbf{W}orld object detection framework (\textbf{IPOW}) based on a Concept Decomposition Model (CDM) from the perspective of concept modeling and decomposition.
% 方法
% Our approach does not rely on mining KUCs. Instead, it grounds unknown detection in reliable semantic concepts learned from annotated known classes, providing stable and interpretable prediction. 
% Each unknown prediction is grounded in activated concepts that explain why it is recognized as unknown. This enables identifying valuable unknowns for annotation and incremental learning, allowing the detector to continuously evolve in real-world scenarios.
Specifically, IPOW is implemented on top of the two-stage detector Faster R-CNN and primarily operates at the RoI head, where each RoI feature is decomposed into three parts: the \textit{discriminative concept}, the \textit{shared concept}, and the \textit{background concept}.
Discriminative concepts are used mainly for the classification of known classes, which are devised to capture the most distinctive attributes within known categories to maximize class separation.
% The Shared concepts capture semantic properties common across known categories and leverage a geometric structure to discover unknown concepts, thereby constructing a comprehensive shared concept space.
To enable the detection of unknown categories, a comprehensive shared concept space is constructed by leveraging an LLM to summarize shared semantic attributes across known categories together with learning complementary shared concepts through feature reconstruction.
Furthermore, through encoding non-object contextual information, background concepts aim to enhance the detection of potential unknown targets via background inversion, \textit{i.e.} identifying regions that deviate from the surrounding context. 
% via encoding non-object contextual information.
% , enabling the detection of potential unknown targets through background inversion, i.e., identifying regions that deviate from the surrounding context.
% Ultimately, decisions for both known and unknown classes can be interpreted by their corresponding concept activations, providing a transparent pathway to demystify the open world detection problem.

% 解释已知未知的混淆
% Through our framework’s interpretable knowledge transfer from known to unknown classes, we demystify the Known-Unknown Confusion.
Drawing on the theory of \textit{Neural Collapse}~\cite{papyan2020prevalence}, as shown in Fig.~\ref{fig:motivation}, known categories collapse into an equiangular tight frame (ETF) structure upon model convergence. 
This ETF structure captures only the most discriminative features to maximize inter-class separation. 
Our discriminative concepts explicitly push known categories into this equiangular structure using the most distinctive features.
Since discriminative concepts are designed solely for known-class recognition, we observe that unknown objects can also fall into this discriminative space. 
As shown in Fig.~\ref{fig:motivation} Right, for example, among known classes such as ``person'' and ``cat,'' the model primarily captures the most distinctive features (two legs vs.\ four legs) to differentiate them. 
However, when encountering a four-legged unknown object such as a ``horse'', the model naturally tends to classify it as a ``cat'', leading to confusion between known and unknown classes.
To resolve this issue, we turn to the shared concept level and propose Concept-Guided Rectification (CGR) based on partial activation of shared concepts.
Because shared concepts are mined for known classes, known objects exhibit ``full activation'' of their corresponding semantic attributes. 
In contrast, unknown objects only trigger ``partial activation'' in this shared space. 
This clear difference in activation patterns allows us to effectively separate unknown objects from known categories. 
% 总结
Overall, this concept-based decomposition not only enables effective separation of unknown categories from known ones, but also provides an explanation of \emph{why unknowns are unknown} at the concept level, thereby realizing interpretable knowledge transfer from known to unknown.

Our contributions are summarized as follows:  
\begin{itemize}
    \item We propose a concept-driven interpretable open-world object detection framework by introducing a Concept Decomposition Model to decompose RoI features into discriminative, shared, and background concepts for known and unknown detection.
    %enabling interpretable results for both known and unknown object detection.

    \item Leveraging this interpretable framework, we identify that known--unknown confusion arises from unknown objects falling into the discriminative space of known classes and propose Concept-Guided Rectification (CGR) to mitigate confusion.

    \item Extensive experiments show that our method achieves state-of-the-art detection performance for known classes and superior recall rate for unknown classes. Moreover, interpretable results for all known and unknown objects can be obtained with the concepts.
\end{itemize}

\section{Related Work}

\begin{figure*}
    \centering
    \includegraphics[width=0.95\linewidth]{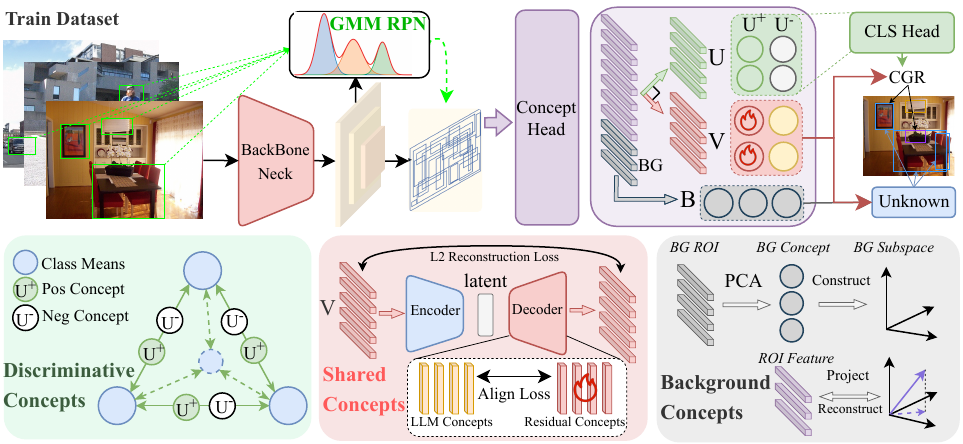}
    \caption{
    Overview of the IPOW framework.  GMM-RPN is utilized to mitigate the proposal generation bias towards known-categories. A concept head is adopted before decomposition. Each RoI feature is decomposed into discriminative, shared, and background concepts for known object recognition, transferable unknown discovery, and contextual modeling. 
    % Concept-guided Rectification (CGR) resolves known--unknown confusion, while the lower modules depict the internal structures of each concept component.
    }
    \label{fig:overallframework}
    \vspace{-3mm}

\end{figure*}

\textbf{Open-World Object Detection.} 
OWOD was first formulated in ORE~\cite{joseph2021towards}, which augments object detection with contrastive clustering and an energy-based unknown classifier. 
OW-DETR~\cite{gupta2022ow} introduces an attention-driven pseudo-labeling scheme for unknown object discovery, while RandBox~\cite{wang2023random} shows that random region proposals significantly improve recall for unknown objects. 
OrthogonalDet~\cite{sun2024exploring} separates objectness detection from classification using orthogonal feature representations, and CROWD~\cite{majee2025looking} uses a submodular-function-based strategy for unknown data discovery and representation learning.
The work most related to ours is a line of attribute-based methods for open-world detection~\cite{xi2025ow, xi2025owovd}, built on pre-trained Open-Vocabulary Object Detectors (OVDs). 
However, OVD-based methods blur the notion of a truly open world: trained on massive image–text corpora, they have effectively \textbf{``seen"} the unknown classes during pre-training, lacking only explicit labels at inference, which causes implicit data leakage.
They also often perform poorly under substantial domain shift from the pre-training data.

\textbf{Concept Bottleneck Models.}  
Concept bottleneck model (CBM)~\cite{koh2020concept} introduces an interpretable framework that predicts human-understandable concepts as intermediate representations before making final predictions. ~\cite{chauhan2023interactive} extend CBM by querying human-provided concept labels at inference time to improve prediction accuracy through concept-level interaction. In addition, semi-supervised CBM~\cite{hu2025semi} incorporate unlabeled data to improve concept learning with limited annotations, while language-guided CBMs~\cite{yu2025language} leverage natural language supervision to align concept representations with semantic knowledge, further boosting interpretability and generalization.
% \textcolor{red}{In this paper, we borrow CBM to reformulate OWOD as an interpretable framework with the help of concept decomposition.}
In this paper, we borrow the CBM and propose CDM to reformulate OWOD as an interpretable framework.
\section{Preliminaries}
\textbf{Problem Definition.}
In line with the conventional OWOD setup, let $\mathcal{T} = \{ \mathcal{T}_1, \dots, \mathcal{T}_n \}$ represent a sequence of tasks arriving over time, each associated with a category set from $\mathcal{C} = \{ \mathcal{C}_1, \dots, \mathcal{C}_n \}$.
At any time step $t$, the system maintains a set of cumulative known categories $\mathcal{K}_t = \{\mathcal{C}_1, \dots, C_t \}$, while the set of unknown classes is explicitly defined as $\mathcal{N}_t = \{ C_{t + 1}, \dots \}$, representing categories that may appear in future tasks. 
The training dataset at time $t$ is denoted as $\mathcal{D}_t = \{(\mathit{I}_i, \mathcal{Y}_i)\}_{i=1}^N$, where $\mathit{I}_i$ denotes the $i_{th}$ image, and 
$\mathcal{Y}_i = \{(\mathbf{b}_k, c_k)\}$ denotes its label containing multiple object instances with bounding boxes $\mathbf{b}_k$ and class labels $c_k$.
% where $\mathcal{Y}_i = \{(\mathbf{b}_k, c_k)\}$ and $c_k \in \mathcal{C}_t$. 
When training task $\mathcal{T}_t$, annotations from previous tasks are unavailable and instances in $\mathcal{N}_t$ are unlabeled.
During inference, the model is required to recognize all categories in $\mathcal{K}_t$ and simultaneously identify instances from $\mathcal{N}_t$ as a generic ``unknown'' class to distinguish them from the background.
As it evolves, the model discovers unknown objects and incorporates them into the known set $\mathcal{K}_{t+1}$ once labels become available, enabling an incremental expansion of known categories while mitigating catastrophic forgetting.

\textbf{Concept Bottleneck Models.}
CBM introduces an interpretable prediction paradigm by explicitly constraining model decisions to pass through a set of human-understandable semantic concepts. Instead of mapping input data to task labels, CBM enforces an intermediate concept space that serves as a bottleneck for reasoning.
Formally, given an input $x \in \mathcal{X}$ and label $y \in \mathcal{Y}$, CBM assumes a concept vector $\vc \in \mathbb{R}^m$, where each dimension corresponds to a concept, and factorizes prediction into two stages:
\begin{equation}
p(y \mid x) = \sum_{c} p(y \mid \vc)\, p(\vc \mid x).
\end{equation}
Here, $p(\vc \mid x)$ models concept prediction, while $p(y \mid \vc)$ performs task inference solely based on concepts.

\section{Method}
% 定义
\subsection{IPOW Framework}
The proposed IPOW is built on the decomposition of RoI features together with semantic concept modeling, which effectively reformulates the OWOD framework for detecting both known and unknown entities.
As shown in Fig.~\ref{fig:overallframework}, IPOW is based on the two-stage detector Faster R-CNN.
% GMM
To eliminate the bias of the RPN towards known categories, we propose a GMM-based RPN for proposal generation; detailed implementation is provided in the Appendix~\ref{appsec:gmm_rpn}.
% 分解
Then through a concept head, the concept feature is decomposed and projected into three concept spaces, namely \textit{Discriminative Concepts}, \textit{Shared Concepts} and \textit{Background Concepts}.
% 作用
Discriminative Concepts are responsible for maximizing inter-class margins among known categories, and the resulting concept activation vectors are fed into a classification head for known-class detection.
Shared Concepts generalize to unknown object detection through LLM-derived concepts and residual concepts learned via the reconstruction process.
Meanwhile, Background Concepts, which model scene context outside object regions, are leveraged to identify regions that are inconsistent with the surrounding background, thereby indicating unknown objects.
Finally, both known and unknown predictions are rectified through the Concept-Guided Rectification (CGR) module to alleviate known–unknown confusion.

\subsection{Concept Decomposition Model}
Given an image $I$, each RoI $x_i$ is encoded as
% Concept head 定义
\begin{equation}
\phi(x_i) = \mathrm{RoIAlign}(\mathrm{FPN}(\mathrm{ResNet}(I)), x_i),
\end{equation}
which is then processed by the Concept Head to produce the concept feature $\mathbf{z}_i \in \mathbb{R}^d$.
\begin{equation}
\mathbf{z}_i = \mathrm{ConceptHead}(\phi(x_i)),
\end{equation}
where the Concept Head consists of a $1 \times 1$ convolution followed by two linear layers. For simplicity, we omit the subscript $i$ in the following.
% dditive
% 前背景分解
Since the feature $\mathbf{z}$ captures both the object of interest and the surrounding background context, we decompose it into two distinct vectors, namely the foreground feature $\mathbf{f}_{\mathrm{fg}}$ and the background feature $\mathbf{f}_{\mathrm{bg}}$, which constitute the foreground and background subspaces, respectively:
\begin{equation}
\mathbf{z} = \mathbf{f}_{\mathrm{fg}} + \mathbf{f}_{\mathrm{bg}}.
\end{equation}
Owing to the mutual exclusivity between foreground and background information, these two components are naturally orthogonal.
% U V 分解
Subsequently, the extracted foreground feature is projected via orthogonal projections \(P_{\mathcal{U}}\) and \(P_{\mathcal{V}}\) into two orthogonal vectors:
\begin{equation}
\mathbf{u} = P_{\mathcal{U}}(\mathbf{f}_{\mathrm{fg}}), \qquad
\mathbf{v} = P_{\mathcal{V}}(\mathbf{f}_{\mathrm{fg}}).
\end{equation}
We define the subspace \(\mathcal{U}\), in which \(\mathbf{u}\) resides, as the \emph{discriminative concept space}, which is utilized to capture the most discriminative features among known classes. In contrast, the subspace \(\mathcal{V}\) is defined as the \emph{shared concept space}, designed to capture semantic attributes shared across known categories, such as having four legs for cats and dogs, or wheels for buses and cars. These shared attributes support generalization for unknown object detection.

The subspaces \(\mathcal{U}\) and \(\mathcal{V}\) are respectively spanned by discriminative concept vectors
$\mathbf{C}^{u} = \{\mathcal{C}^{u}_{1}, \ldots, \mathcal{C}^{u}_{K_u} \}$
 and shared concept vectors
\(\mathbf{C}^{v} = \{ \mathcal{C}^{v}_{1}, \ldots, \mathcal{C}^{v}_{K_v} \}\).
To enforce strict orthogonality, we construct the two subspaces by projecting the original concept vectors using two fixed, mutually orthogonal \emph{orthogonal matrices} \(\mathbf{Q}_{\mathcal{U}}\) and \(\mathbf{Q}_{\mathcal{V}}\):
\begin{equation}
\mathcal{U} = \mathrm{span}\big( \mathbf{Q}_{\mathcal{U}} \, \mathbf{C}^{u} \big), \qquad
\mathcal{V} = \mathrm{span}\big( \mathbf{Q}_{\mathcal{V}} \, \mathbf{C}^{v} \big),
\end{equation}
Accordingly, the foreground space is defined as
\begin{equation}
\mathcal{Z}_{\mathrm{fg}} = \mathcal{U} \oplus \mathcal{V}, \qquad \mathcal{U} \perp \mathcal{V},\qquad 
\end{equation}
and the foreground representation is expressed as
\begin{equation}
\mathbf{f}_{\mathrm{fg}} = \mathbf{u} + \mathbf{v}.
\end{equation}
By substituting the foreground decomposition, the complete formulation of the concept feature is expressed as:
\begin{equation}
    \mathbf{z} = \mathbf{u} + \mathbf{v} + \mathbf{f}_{\mathrm{bg}}.
\end{equation}
Consequently, an unknown instance can be represented as:
\begin{equation}
    \mathbf{z}^{\mathrm{unk}} = \mathbf{u}^{\mathrm{unk}} + \mathbf{v}^{\mathrm{unk}} + \mathbf{f}_{\mathrm{bg}}.
\end{equation}
In OWOD, unknown instances need not be explicitly classified, so $\mathbf{u}^{\mathrm{unk}}$ is not modeled and the core lies in $\mathbf{v}^{\mathrm{unk}}$. Given the commonalities across object categories, $\mathbf{v}^{\mathrm{unk}}$ and $\mathbf{v}^{\mathrm{known}}$ are expected to be highly overlapping within the shared subspace. Therefore, the model can effectively generalize to unknown categories by modeling the shared concepts derived from known classes. 
Furthermore, as background encodes only category-agnostic scene context invariant across known and unknown data, explicitly modeling $\mathbf{f}_{\mathrm{bg}}$ facilitates unknown detection by distinguishing foreground from the environment.
The details of each concept submodule are as follows.

\subsection{Discriminative Concepts}
% NC
According to the Neural Collapse theory~\cite{papyan2020prevalence}, when a classifier is trained to convergence on a set of known classes, the feature means of different classes collapse into an \emph{Equiangular Tight Frame} (ETF), referred to as NC2. Formally, let $\{\boldsymbol{\mu}_k\}_{k=1}^K$ denote the class-wise feature means. Neural Collapse implies that
\begin{equation}
\begin{aligned}
\|\boldsymbol{\mu}_k\|_2 &= \|\boldsymbol{\mu}_{k'}\|_2, \\
\langle \boldsymbol{\mu}_k, \boldsymbol{\mu}_{k'} \rangle &= -\frac{1}{K-1}, \quad \forall k \neq k'.
\end{aligned}
\end{equation}
% These results indicate that deep networks learn maximally separable features among classes.

Motivated by this observation, we define \emph{discriminative concepts} as the most discriminative positive--negative concept pairs between every two known categories, thereby driving known class representations toward an ETF structure, as illustrated in Fig.~\ref{fig:overallframework}. Specifically, we leverage an LLM to identify the most discriminative attributes between two classes. Concepts possessing the attribute are treated as positive concepts, while those lacking it are treated as negative concepts, forming a set of discriminative concept pairs $\{(\mathcal{C}^{u+}_{i}, \mathcal{C}^{u-}_{i})\}$.
Each concept $\mathcal{C}^{u}_{i}$ is encoded using the CLIP text encoder to obtain a concept embedding. The activation of concept $\mathcal{C}^{u}_{i}$ with respect to the discriminative feature vector $\mathbf{u}$ is computed via cosine similarity:
\begin{equation}
    \mathbf{e}^{u}_i = \mathrm{TextEnc}(\mathcal{C}^{u}_{i}), \quad \quad a(\mathcal{C}^{u}_{i}) = \frac{\mathbf{u}^\top \mathbf{e}^{u}_i}{\|\mathbf{u}\|_2 \, \|\mathbf{e}^{u}_i\|_2}.
\end{equation}
Accordingly, the discriminative concept space is optimized using a margin-based contrastive objective:
\begin{equation}
\mathcal{L}_{\mathrm{disc}} = \sum_{i} \Big[ \|a(\mathcal{C}^{u+}_{i}) - a(\mathcal{C}^{u-}_{i})\|_2^2 - \delta \Big] ,
\end{equation}
where $\delta > 0$ is a margin.
Following the standard CBM architecture, we attach a linear classification head to the discriminative concept activations to produce the final classification results :
\begin{equation}
\label{eq:dis_cls_score}
S_{\mathrm{cls}}^k = p(y = k \mid \mathbf{z}) = \mathrm{Softmax}\!\left(\mathbf{W}_k^\top \mathbf{a}^{u}\right),
\end{equation}
where $\mathbf{a}^{u} = [a(\mathcal{C}^{u}_{1}), \ldots, a(\mathcal{C}^{u}_{K_u})]^\top$ denotes the vector of activations over all $K_u$ discriminative concepts.
such that known-class predictions are determined solely by discriminative concept activations. 
% This design yields compact intra-class representations and maximal inter-class separation, while enabling direct interpretability through concept activations.

\subsection{Shared Concepts}
% LLM 总结已知概念
The shared concept space $\mathcal{V}$ is designed to explicitly construct a comprehensive set of semantic attributes that are shared across known classes. 
To achieve this, we first summarize shared concepts from known categories using an LLM, producing a set of human-interpretable shared concepts $\{\mathcal{C}^{v}_{i}\}_{i=1}^{K}$. 
Each shared concept $\mathcal{C}^{v}_{i}$ is encoded using the CLIP text encoder, and its activation with respect to the shared feature vector $\mathbf{v}$ is computed via cosine similarity:
\begin{equation}
    \mathbf{e}^{v}_i = \mathrm{TextEnc}(\mathcal{C}^{v}_{i}), \quad a(\mathcal{C}^{v}_{i}) = \frac{\mathbf{v}^\top \mathbf{e}^{v}_i}{\|\mathbf{v}\|_2 \, \|\mathbf{e}^{v}_i\|_2}.
\end{equation}
The shared representation  optimized via binary cross-entropy supervision,
for a given category, the shared concept learning objective is formulated as:
\begin{equation}
\mathcal{L}_{\mathrm{sc}} = - \sum_{i} 
\Big[
y_i \log(a(\mathcal{C}^{v}_{i})) + (1 - y_i)\log(1 - a(\mathcal{C}^{v}_{i}))
\Big],
\end{equation}
where $y_i \in \{0,1\}$ indicates the presence of the $i$-th shared concept for the category, and $a(\mathcal{C}^{v}_{i})$ denotes the predicted activation of the corresponding shared concept.

% OWOD 实验结果
\begin{table*}[t]
\centering
\caption{OWOD performance comparison against state-of-the-art approaches on M-OWODB (top) and S-OWODB (bottom).
We report U-Recall for unknown classes and mAP for known classes.
Best results are shown in bold.
For a fair comparison, * indicates results reproduced under the setting of limiting the number of unknown proposals to 100 per image. Best results in bold.}
\label{tab:owod_results}
% 表头
\adjustbox{width=\textwidth}{
\renewcommand{\arraystretch}{1.2} % 稍微增加行间距，使表格不拥挤
\begin{tabular}{c|cc|cccc|cccc|ccc}
\toprule
\textbf{Task IDs(→)} & \multicolumn{2}{c|}{\textbf{Task 1}} & \multicolumn{4}{c|}{\textbf{Task 2}} & \multicolumn{4}{c|}{\textbf{Task 3}} & \multicolumn{3}{c}{\textbf{Task 4}} \\
\hline
\multirow{2}{*}{\textbf{Methods}} & \cellcolor{lightyellow} & \cellcolor{lightblue} mAP & \cellcolor{lightyellow} & \multicolumn{3}{c|}{\cellcolor{lightblue} mAP} & \cellcolor{lightyellow} & \multicolumn{3}{c|}{\cellcolor{lightblue} mAP} & \multicolumn{3}{c}{\cellcolor{lightblue} mAP} \\
 & \multirow{-2}{*}{\cellcolor{lightyellow} U-Recall} & Curr. & \multirow{-2}{*}{\cellcolor{lightyellow} U-Recall} & Prev. & Curr. & Both & \multirow{-2}{*}{\cellcolor{lightyellow} U-Recall} & Prev. & Curr. & Both & Prev. & Curr. & Both \\
\hline

% M-OWODB 实验表格
\multicolumn{14}{c}{\textbf{M-OWODB}} \\
\hline
ORE~\cite{joseph2021towards} & \cellcolor{lightyellow} 4.9 & 56.0 & \cellcolor{lightyellow} 2.9 & 52.7 & 26.0 & 39.4 & \cellcolor{lightyellow} 3.9 & 38.2 & 12.7 & 29.7 & 29.6 & 12.4 & 25.3 \\
OW-DETR~\cite{gupta2022ow} & \cellcolor{lightyellow} 7.5 & 59.2 & \cellcolor{lightyellow} 6.2 & 53.6 & 33.5 & 42.9 & \cellcolor{lightyellow} 5.7 & 38.3 & 15.8 & 30.8 & 31.4 & 17.1 & 27.8 \\
PROB~\cite{zohar2023prob} & \cellcolor{lightyellow} 19.4 & 59.5 & \cellcolor{lightyellow} 17.4 & 55.7 & 32.2 & 44.0 & \cellcolor{lightyellow} 19.6 & 43.0 & 22.2 & 36.0 & 35.7 & 18.9 & 31.5 \\
CAT~\cite{ma2023cat} & \cellcolor{lightyellow} 23.7 & 60.0 & \cellcolor{lightyellow} 19.1 & 55.5 & 32.2 & 44.1 & \cellcolor{lightyellow} 24.4 & 42.8 & 18.8 & 34.8 & 34.4 & 16.6 & 29.9 \\
RandBox~\cite{wang2023random} & \cellcolor{lightyellow} 10.6 & 61.8 & \cellcolor{lightyellow} 6.3 & - & - & 45.3 & \cellcolor{lightyellow} 7.8 & - & - & 39.4 & - & - & 35.4 \\
OrthogonalDet~\cite{sun2024exploring} & \cellcolor{lightyellow} 24.6 & 61.3 & \cellcolor{lightyellow} 26.3 & 55.5 & 38.5 & 47.0 & \cellcolor{lightyellow} 29.1 & 46.7 & 30.6 & 41.3 & 42.4 & 24.3 & 37.9 \\
$\mathrm{CROWD^*}$~\cite{majee2025looking} & \cellcolor{lightyellow} 42.9 & 61.7 & \cellcolor{lightyellow} 31.4 & 56.7 & 38.9 & 47.8 & \cellcolor{lightyellow} 34.7 & 48.0 & 31.4 & 42.5 & 42.9 & 25.4 & 38.5 \\
\hline
\textbf{IPOW (Ours)} & \cellcolor{lightyellow} \textbf{50.1} & \textbf{62.4} & \cellcolor{lightyellow} \textbf{41.9} & \textbf{61.7} & \textbf{43.6} & \textbf{52.7} & \cellcolor{lightyellow} \textbf{46.3} & \textbf{49.7} & \textbf{35.5} & \textbf{45.0} & \textbf{46.7} & \textbf{28.6} & \textbf{42.2} \\
\hline

% S-OWODB
\multicolumn{14}{c}{\textbf{S-OWODB}} \\
\hline
ORE~\cite{joseph2021towards} & \cellcolor{lightyellow} 1.5 & 61.4 & \cellcolor{lightyellow} 3.9 & 56.7 & 26.1 & 40.6 & \cellcolor{lightyellow} 3.6 & 38.7 & 23.7 & 33.7 & 33.6 & 26.3 & 31.8 \\
OW-DETR~\cite{gupta2022ow} & \cellcolor{lightyellow} 5.7 & 71.5 & \cellcolor{lightyellow} 6.2 & 62.8 & 27.5 & 43.8 & \cellcolor{lightyellow} 6.9 & 45.2 & 24.9 & 38.5 & 38.2 & 28.1 & 33.1 \\
PROB~\cite{zohar2023prob} & \cellcolor{lightyellow} 17.6 & 73.4 & \cellcolor{lightyellow} 22.3 & 66.3 & 36.0 & 50.4 & \cellcolor{lightyellow} 24.8 & 47.8 & 30.4 & 42.0 & 42.6 & 31.7 & 39.9 \\
CAT~\cite{ma2023cat} & \cellcolor{lightyellow} 24.0 & 74.2 & \cellcolor{lightyellow} 23.0 & 67.6 & 35.5 & 50.7 & \cellcolor{lightyellow} 24.6 & 51.2 & 32.6 & 45.0 & 45.4 & 35.1 & 42.8 \\
OrthogonalDet~\cite{sun2024exploring} & \cellcolor{lightyellow} 24.6 & 71.6 & \cellcolor{lightyellow} 27.9 & 64.0 & 39.9 & 51.3 & \cellcolor{lightyellow} 31.9 & 52.1 & 42.2 & 48.8 & 48.7 & 38.8 & 46.2 \\
$\mathrm{CROWD^*}$~\cite{majee2025looking} & \cellcolor{lightyellow} 30.4 & 73.5 & \cellcolor{lightyellow} 25.5 & 64.9 & 41.2 & 53.1 & \cellcolor{lightyellow} 38.1 & 54.7 & 42.1 & 48.4 & 49.8 & 43.0 & 46.4 \\
\hline
\textbf{IPOW (Ours)} & \cellcolor{lightyellow} \textbf{34.7} & \textbf{73.6} & \cellcolor{lightyellow} \textbf{32.6} & \textbf{67.9} & \textbf{42.3} & \textbf{55.1} & \cellcolor{lightyellow} \textbf{44.3} & \textbf{56.3} & \textbf{44.6} & \textbf{50.5} & \textbf{50.2} & \textbf{44.6} & \textbf{47.4} \\
\bottomrule
\end{tabular}
} % 
\vspace{-3mm}
\end{table*}

% 未知概念
However, LLM-derived concepts are inherently incomplete and cannot exhaustively characterize all transferable semantics. To address this limitation, as shown in Fig.\ref{fig:overallframework} we employ a sparse auto-encoding mechanism to discover residual shared concepts beyond those summarized by LLM. Given a shared feature $\mathbf{v} \in \mathcal{V}$, we first apply a linear encoder to project it into a low-dimensional sparse activation vector:
\begin{equation}
\mathbf{\alpha} = \mathrm{Enc}(\mathbf{v}) = \mathbf{W_e} \mathbf{v}, \qquad \mathbf{\alpha} \in \mathbb{R}^m,
\end{equation}
where $\mathbf{W_e} \in \mathbb{R}^{m \times d}$ and sparsity is enforced on $\alpha$. The decoder consists of a concept dictionary composed of both known and learnable residual shared concept vectors:
\begin{equation}
\mathbf{D} = \big[ \mathbf{D^{\mathrm{k}}},\;\mathbf{D^{\mathrm{r}}} \big] \in \mathbb{R}^{d \times m},
\end{equation}
where $\mathbf{D}^{\mathrm{k}} = \{\mathbf{e}^{v}_i\}_{i=1}^{K}$ corresponds to the LLM-derived shared concepts, and $\mathbf{D}^{\mathrm{r}} = [\mathbf{d}^{\mathrm{r}}_1, \ldots, \mathbf{d}^{\mathrm{r}}_M]$ consists of $M$ randomly initialized, learnable shared concept vectors.
The shared feature is reconstructed as
\begin{equation}
\hat{\mathbf{v}} = \mathbf{D} \alpha, \quad
\mathcal{L}_{\mathrm{rec}} = \| \mathbf{v} - \hat{\mathbf{v}} \|_2^2,
\end{equation}
Finally, to ensure that the residual shared concepts are effectively activated and complementary to the LMM-derived ones, we apply the following regularization loss:
\begin{equation}
\mathcal{L}_{\mathrm{align}}
=
\left\|
(\mathbf{D}^{\mathrm{k}})^{\top}\mathbf{D}^{\mathrm{r}}
\right\|_{F}^{2}
+ 
\left|
\mathcal{E}_{\ell_2}\!\left(\mathbf{D}^{\mathrm{k}}\right)
-
\mathcal{E}_{\ell_2}\!\left(\mathbf{D}^{\mathrm{r}}\right)
\right| .
\end{equation}
Here, $\mathcal{E}_{\ell_2}(\cdot)$ denotes the $\ell_2$ activation energy of a shared concept vector $\mathbf{v}$ over the corresponding concept group. 

Accordingly, the LLM-derived and residual shared concepts are unified into a complete shared concept set
$\{\mathcal{C}^{v}_{i}\}_{i=1}^{K+M}$, with corresponding concept activations $a(\mathcal{C}^{v}_{i})$.
The unknownness score is then defined as the maximum activation over the shared concept set: $S_{\mathrm{unk}}^{\mathrm{share}} = \max_{i} \; a(\mathcal{C}^{v}_{i}).$

\subsection{Background Concepts}

Background Concepts are obtained by applying principal component analysis (PCA) to a set of Background ROI features, producing a set of orthonormal basis vectors.  We define these basis vectors as Background Concepts:
\begin{equation}
\mathbf{D}_{\mathrm{bg}} = [\mathcal{C}^{b}_{1}, \mathcal{C}^{b}_{2}, \dots, \mathcal{C}^{b}_{k} \in \mathbb{R}^{d \times k}],
\end{equation}
which span the Background Subspace $\mathcal{F}_{\mathrm{bg}} = \mathrm{span}(\mathbf{D_{\mathrm{bg}}})$. Given a RoI Feature $\textbf{z} \in \mathbb{R}^d$, we compute its reconstruction from the Background Subspace by projecting onto this basis, 
\begin{equation}
\hat{\mathbf{z}} = \mathbf{D_{\mathrm{bg}}} \mathbf{D_{\mathrm{bg}}^\top} \mathbf{z}, \quad r(\textbf{z}) = \| \textbf{z} - \hat{\textbf{z}} \|_2
\end{equation}
The reconstruction error $r(\textbf{z})$ measures how well $\textbf{z}$ can be explained by Background Concepts. A larger error indicates a feature inconsistency with background patterns, i.e., likely belonging to a foreground object, possibly an unknown class.
Thus, the foreground score is defined as a normalized reconstruction error,
$S_{\mathrm{unk}}^{\mathrm{bg}} = \mathrm{Norm}\!\left(r(\mathbf{z})\right)$ with respect to the feature magnitude, serving as a category-agnostic indicator for unknown object detection.

\subsection{Concept Guided Rectification}
% “概念引导修正”机制
% 4.5 模型推理 这里要有利用概念的矫正 未知目标是部分概念激活
% 但已知目标要所有概念激活
% By reconstructing the knowledge transfer process in OWOD through concepts, we identify the root cause of known–unknown confusion.
Our core insight is that known objects must exhibit full-set activation of their predefined semantic concepts, whereas unknown objects may fall into the discriminative space $\mathcal{U}$ of known categories but typically trigger only partial activation within the shared space $\mathcal{V}$.
Based on this insight, we introduce a concept-guided rectification mechanism. 
% back up lvxueqiang
% For a candidate RoI $x_i$, let $\hat{\mathbf{c}} = \{\hat{c}_k\}_{k=1}^{K_s}$ denote the predicted activations for shared concepts. For each known class $j$ with its associated shared concept set $\mathcal{C}_j$, the rectified confidence score $S_{\mathrm{known}}^j$ is defined as:
% \begin{equation}
% S_{\mathrm{known}}^j
% =
% S_{\mathrm{cls}}^j \cdot
% \left(
% \prod_{c_k \in \mathcal{C}_j} \hat{c}_k
% \right)^{\frac{\eta}{|\mathcal{C}_j|}},
% \end{equation}
For a candidate RoI $x_i$, let $\hat c_k \in [0,1]$ denote the predicted activation of the $k$-th shared concept. For each known class $j$ with its associated shared concept set $\mathcal{C}_j$, and let $|\mathcal{C}_j|$ denote its cardinality.
The rectified confidence score $S_{\mathrm{known}}^j$ is defined as:
\begin{equation}
S_{\mathrm{known}}^j
=
S_{\mathrm{cls}}^j \cdot
\left(
\prod_{c_k \in \mathcal{C}_j} \hat{c}_k
\right)^{\frac{\eta}{|\mathcal{C}_j|}},
\end{equation}

where \(S_{\mathrm{cls}}^j\) denotes the raw classification probability obtained by Eqn.~\ref{eq:dis_cls_score} and \(\eta\) is a scaling factor that controls the strength of concept-guided rectification.
% This multiplicative constraint enforces strict conceptual consistency, ensuring that a high score is maintained only when all the requisite semantic attributes are present. 
Conversely, unknown objects are identified by shared and background concept activations that fail to satisfy the strict criteria of any known class. 
The rectified unknown score is defined as
\begin{equation}
S_{\mathrm{unk}} =
\max\!\left(
S_{\mathrm{unk}}^{\mathrm{share}},
S_{\mathrm{unk}}^{\mathrm{bg}}
\right)
\cdot
\left(1 - \max_{j} S_{\mathrm{known}}^{j}\right),
\end{equation}

Ultimately, known-category inference is driven by discriminative concepts and rectified through shared concepts, while unknown-category inference is determined by joint activations of shared and background concepts. 
As a result, all predictions are grounded in concept-level evidence, leading to robust and interpretable open-world object detection.

\section{Experiments}
% Unknown object confusion on M-OWODB
\begin{table*}[t]
    \centering
    \caption{Known–Unknown confusion on M-OWODB. The comparison is shown in terms of U-Recall, WI and A-OSE. Note that these metrics are not calculated for Task 4 because all 80 classes are known. Best results in bold.}
    \label{tab:m_owodb_unknown}
    
    \begin{adjustbox}{width=\textwidth}
        \renewcommand{\arraystretch}{1.2}
        \setlength{\tabcolsep}{8pt}
        \begin{tabular}{l|ccc|ccc|ccc}
            \toprule
            \textbf{Task IDs ($\rightarrow$)} & \multicolumn{3}{c|}{\textbf{Task 1}} & \multicolumn{3}{c|}{\textbf{Task 2}} & \multicolumn{3}{c}{\textbf{Task 3}} \\
            \cmidrule{2-10}
            \textbf{Method} & \cellcolor{lightyellow}U-Recall ($\uparrow$) & \cellcolor{lightblue}WI ($\downarrow$) & \cellcolor{lightblue}A-OSE ($\downarrow$) & \cellcolor{lightyellow}U-Recall ($\uparrow$) & \cellcolor{lightblue}WI ($\downarrow$) & \cellcolor{lightblue}A-OSE ($\downarrow$) & \cellcolor{lightyellow}U-Recall ($\uparrow$) & \cellcolor{lightblue}WI ($\downarrow$) & \cellcolor{lightblue}A-OSE ($\downarrow$) \\
            \midrule
            ORE \cite{joseph2021towards}         & \cellcolor{lightyellow}4.9  & 0.0621 & 10459 & \cellcolor{lightyellow}2.9  & 0.0282 & 10445 & \cellcolor{lightyellow}3.9  & 0.0211 & 7990 \\
            OW-DETR \cite{gupta2022ow}  & \cellcolor{lightyellow}7.5  & 0.0571 & 10240 & \cellcolor{lightyellow}6.2  & 0.0278 & 8441  & \cellcolor{lightyellow}5.7  & 0.0156 & 6803 \\
            PROB \cite{zohar2023prob}       & \cellcolor{lightyellow}19.4 & 0.0569 & 5195  & \cellcolor{lightyellow}17.4 & 0.0344 & 6452  & \cellcolor{lightyellow}19.6 & 0.0151 & 2641 \\
            RandBox \cite{wang2023random} & \cellcolor{lightyellow}10.6 & \textbf{0.0240} & 4498 & \cellcolor{lightyellow}6.3  & \textbf{0.0078} & 1880  & \cellcolor{lightyellow}7.8  & \textbf{0.0054} & 1452 \\
            
            OrthogonalDet \cite{sun2024exploring} & \cellcolor{lightyellow}24.6 & 0.0299 & 4148 & \cellcolor{lightyellow}26.3  & 0.0099 & 1791  & \cellcolor{lightyellow} 29.1  & 0.0077 & 1345 \\
            
            $\mathrm{CROWD^*}$ \cite{majee2025looking} & \cellcolor{lightyellow}42.9 & 0.0380 & 3823 & \cellcolor{lightyellow}31.4  & 0.0101 & 1508  & \cellcolor{lightyellow}34.7  & 0.0066 & 1266 \\
            \midrule
            
            \textbf{Ours}          & \cellcolor{lightyellow}\textbf{50.1} & 0.0369 & \textbf{3648} & \cellcolor{lightyellow}\textbf{41.9} & 0.0098 & \textbf{1460} & \cellcolor{lightyellow}\textbf{46.3} & 0.0062 & \textbf{1126} \\
            \bottomrule
        \end{tabular}
    \end{adjustbox}
\end{table*}

% Ablation Study
\begin{table*}[t]
    \centering
\caption{The ablation study of each component on the M-OWODB benchmark across Tasks~1–3. Best results in bold.}
% We evaluate the contribution of each IPOW component using unknown recall (U-Recall), known-class mAP, wilderness impact (WI), and absolute open-set error (A-OSE).}
    \label{tab:ablation_mowodb}
    \begin{adjustbox}{width=\textwidth}
        \renewcommand{\arraystretch}{1.3}
        \begin{tabular}{l|cccc|cccc|cccc}
            \toprule
            \textbf{Task IDs ($\rightarrow$)} & \multicolumn{4}{c|}{\textbf{Task 1}} & \multicolumn{4}{c|}{\textbf{Task 2}} & \multicolumn{4}{c}{\textbf{Task 3}} \\
            \hline
            \multirow{2}{*}{\textbf{Methods}} & \cellcolor{lightyellow} U-Recall & \cellcolor{lightyellow} mAP & \cellcolor{lightblue} WI & \cellcolor{lightblue} A-OSE & \cellcolor{lightyellow} U-Recall & \cellcolor{lightyellow} mAP & \cellcolor{lightblue} WI & \cellcolor{lightblue} A-OSE & \cellcolor{lightyellow} U-Recall & \cellcolor{lightyellow} mAP & \cellcolor{lightblue} WI & \cellcolor{lightblue} A-OSE \\
             & \cellcolor{lightyellow} ($\uparrow$) & \cellcolor{lightyellow} ($\uparrow$) & \cellcolor{lightblue} ($\downarrow$) & \cellcolor{lightblue} ($\downarrow$) & \cellcolor{lightyellow} ($\uparrow$) & \cellcolor{lightyellow} ($\uparrow$) & \cellcolor{lightblue} ($\downarrow$) & \cellcolor{lightblue} ($\downarrow$) & \cellcolor{lightyellow} ($\uparrow$) & \cellcolor{lightyellow} ($\uparrow$) & \cellcolor{lightblue} ($\downarrow$) & \cellcolor{lightblue} ($\downarrow$) \\
            \hline
            Base Model & 21.8 & 61.1 & 0.0566 & 6765 & 14.7 & 51.9 & 0.0268 & 2543 & 17.7 & 43.8 & 0.0151 & 1676 \\
            + GMM RPN & 25.9 & 61.7 & 0.0536 & 8029 & 19.1 & 52.1 & 0.0269 & 3897 & 22.4 & 43.6 & 0.0164 & 2670 \\
            + Discriminative Concepts & 23.3 & 62.4 & 0.0492 & 6778 & 19.3 & 52.4 & 0.0273 & 3721 & 21.4 & 44.2 & 0.0173 & 2720 \\
            + Share Concept & 45.4 & 62.6 & 0.0460 & 6264 & 37.6 & 52.5 & 0.0286 & 3746 & 40.5 & 44.7 & 0.0156 & 2650 \\
            + BG Concept & \textbf{50.4} & \textbf{62.6} & 0.0460 & 6264 & \textbf{42.4} & 52.5 & 0.0286 & 3746 & \textbf{46.9} & 44.7 & 0.0156 & 2650 \\
            + Concept-Guided Rectification & 50.1 &  62.4 & \textbf{0.0369} & \textbf{3648} & 41.9 &  \textbf{52.7} & \textbf{0.0098} & \textbf{1460} & 46.3 &  \textbf{45.0} & \textbf{0.0062} & \textbf{1126} \\
            \bottomrule
        \end{tabular}
    \end{adjustbox}
    \vspace{-3mm}
\end{table*}

% DIOR 实验结果
\begin{table}[t]
\centering
\caption{OWOD results on the Remote Sensing DIOR dataset.
% U-Recall, known-class mAP,  and confusion metrics (WI and A-OSE) are reported.
}
\label{tab:owod_dior_recall}
\begin{adjustbox}{width=1\linewidth}
\renewcommand{\arraystretch}{1.3}

\begin{tabular}{c|cccc|ccc}
\toprule
\textbf{Task IDs} 
& \multicolumn{4}{c|}{\textbf{Task 1}} 
& \multicolumn{3}{c}{\textbf{Task 2}} \\
\hline
\multirow{2}{*}{\textbf{Methods}} 
& \cellcolor{lightyellow}
& \cellcolor{lightblue}mAP 
& \cellcolor{lightblue}
& \cellcolor{lightblue}
& \multicolumn{3}{c}{\cellcolor{lightblue}mAP} \\
& \multirow{-2}{*}{\cellcolor{lightyellow} U-Recall}
& \cellcolor{lightblue}Curr. 
& \multirow{-2}{*}{\cellcolor{lightblue}WI} 
& \multirow{-2}{*}{\cellcolor{lightblue}A-OSE} 
& \cellcolor{lightblue}Prev. 
& \cellcolor{lightblue}Curr. 
& \cellcolor{lightblue}Both \\
\hline
Base Model 
& 2.8 & 69.8 & 0.0743 & 1833 & 67.0 & 64.3 & 65.6 \\
IPOW 
& \textbf{19.9} & \textbf{71.2} & \textbf{0.0573} & \textbf{1143} 
& \textbf{68.2} & \textbf{66.3} & \textbf{67.2} \\
\bottomrule
\end{tabular}

\end{adjustbox}

\vspace{-4mm}

\end{table}

\subsection{Experiment Setup} 
Details about the dataset and implementation can be found in Appendix~\ref{appsec:experiment_details}.

\paragraph{Evaluation Metrics.}  
We evaluate known classes using mean Average Precision (mAP), reported separately for previously seen and newly introduced categories. 
For unknown object classes, we follow the standard OWOD protocol~\cite{joseph2021towards, sun2024exploring} and report Unknown Object Recall (U-Recall). 
Since U-Recall can be trivially increased by predicting an excessive number of unknown proposals, it may lose its practical significance.
We improve the U-Recall metric by limiting the number of post-NMS proposals to at most 100 per image, consistent with the standard Faster R-CNN setting.
This refinement enables a fair comparison between different methods in terms of their performance on unknown detection.
To quantify confusion between known and unknown classes, we also report Wilderness Impact (WI)~\cite{dhamija2020overlooked} and Absolute Open-Set Error (A-OSE)~\cite{miller2018dropout}.

\subsection{Performance on OWOD}
% \textbf{Results on M-OWODB.} As shown in Table~\ref{tab:owod_results} top, IPOW achieves the best overall performance on M-OWODB across all incremental tasks.
% Compared with the strongest 之前方法 CROWD, IPOW improves unknown recall by +7.5, +14.4 and +13.6 on Tasks~1–3,  IPOW attains the highest unknown recall on all tasks (50.4\%, 45.8\%, 48.3\%, and 44.2\%) and consistently achieves the bast known-class mAP  这里列出数据对比。
% 根据结果可以看出, IPOW随之任务增加 已知类别增多，学习到的概念变得更丰富，未知的recall的优势会明显增加，这个是IPOW的独特优势，也是真实世界中明确要求
\textbf{Results on M-OWODB.}
As shown in the upper part of Table~\ref{tab:owod_results}, IPOW achieves the best overall performance on M-OWODB.
Compared to previous state-of-the-art method CROWD, IPOW improves U-Recall by \textit{7.2}, \textit{10.5}, and \textit{11.6} points on Tasks~1–3, respectively, achieving remarkable U-Recall rates of 50.1\%, 41.9\% and 46.3\%.
Meanwhile, IPOW also achieves the best known-class mAP across tasks, with mAP values of 62.4, 52.7, 45.0, and 42.2 on both classes.
As more known categories are introduced, IPOW achieves larger U-Recall gains over existing methods, suggesting increasingly rich and transferable concepts consistent with open-world detection practical settings.

% \textbf{Results on S-OWODB.}
% 在 superclass-separated S-OWODB As shown in Table~\ref{tab:owod_results} bottom, IPOW consistently achieves the best performance。
% Compared with the strongest baseline CROWD, IPOW improves unknown recall by +4.3, +7.1, +6.2 points on Tasks~1–3 attains the highest unknown recall on all tasks (34.7\%, 32.6\% and 44.3\%) and consistently improves the joint mAP 达到..
% 对比来说 S-OWODB 对于 IPOW会更难，因为任务按照超类进行分离导致可以进行泛化的共享概念变得更少，但 IPOW 仍然表现出优于之前方法的结果，证明基于概念的知识迁移是正确的路径。

\textbf{Results on S-OWODB.}
On the superclass-separated S-OWODB benchmark, IPOW consistently achieves the best performances, as shown in the lower part of Table~\ref{tab:owod_results}.
Compared to CROWD, IPOW improves U-Recall by \textit{4.3}, \textit{7.1}, and \textit{6.2} points in Tasks~1–3, respectively, achieving the highest U-Recall in all tasks (34.7\%, 32.6\% and 44.3\%).
Meanwhile, IPOW also consistently improves the mAP of known categories, reaching 73.6, 55.1, 50.5, and 47.4 on Tasks~1–4, respectively. 
Note that S-OWODB is more challenging for our IPOW, as superclass-level separation reduces the number of transferable shared concepts.
Despite this difficulty, IPOW can still significantly outperform previous methods, showing that concept decomposition model is an effective solution for OWOD.

%  back up for 1.28 night
% \textbf{Known--Unknown Confusion Analysis.}
% \textcolor{red}{We further analyze known--unknown confusion in Table~\ref{tab:m_owodb_unknown}.
% Across all tasks, IPOW consistently achieves the lowest A-OSE, reducing false known predictions to 3648, 1460, and 1126, respectively, Compared to CROWD, it decreased by 175, 48, 140 respectively. while maintaining very competitive Wilderness Impact (WI) values of 0.0369, 0.0098, and 0.0062.

% These results provide clear evidence that our analysis of the cause of known--unknown confusion is valid, 
% and that the proposed concept-guided rectification mechanism is effective in mitigating such confusion.
% }
\textbf{Known--Unknown Confusion Analysis.}
We further analyze known--unknown confusion in Table~\ref{tab:m_owodb_unknown}.
Across all tasks, IPOW consistently achieves the lowest A-OSE, reducing false known predictions to 3648, 1460, and 1126, respectively.
Compared to CROWD, this corresponds to reductions of 175, 48, and 140 unknown objects being wrongly classified as any of the known class.
Meanwhile, IPOW maintains highly competitive Wilderness Impact (WI) values of 0.0369, 0.0098, and 0.0062.
These results clearly show that our method has a great advantage in reducing known-unknown confusion and the proposed concept-guided rectification mechanism is effective in mitigating such confusion.

\textbf{Results on the Remote Sensing Dataset DIOR.}
As shown in Table~\ref{tab:owod_dior_recall}, IPOW significantly improves U-Recall on DIOR, increasing U-Recall from 2.8\% to 19.9\% on Task~1, and greatly reduces known--unknown confusion, decreasing WI from 0.0743 to 0.0573 and A-OSE from 1833 to 1143.
In addition, for known classes, IPOW improves mAP by 1.4 and 1.6 points in Task~1--2, respectively.
These results show that IPOW generalizes well to remote sensing scenarios that differ significantly from everyday natural images.

\begin{figure}
	\centering
	\includegraphics[width=1\linewidth]{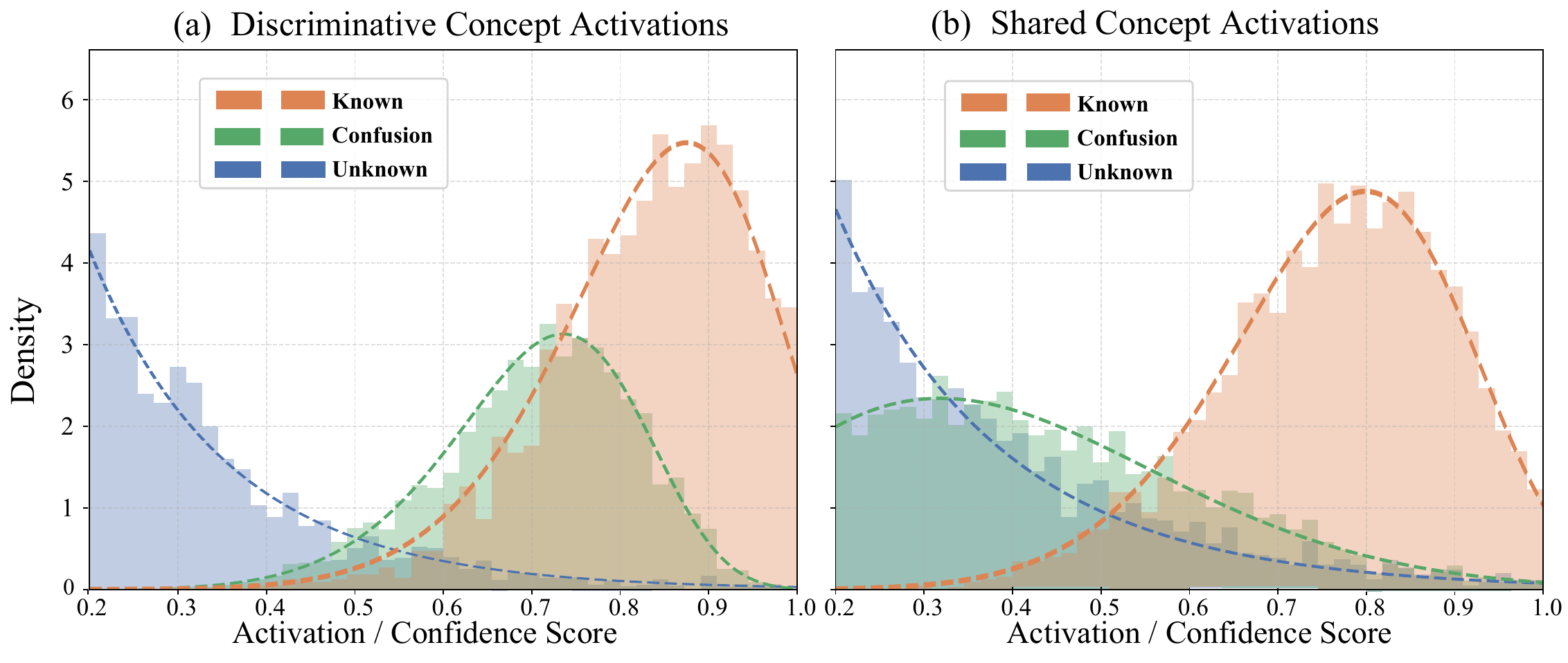}
\caption{Score distributions on M-OWODB Task 1 for known, confusion (unknown falsely predicted as known), and unknown samples. Discriminative concept activations (left) and the geometric mean of class-specific shared concept activations (right).}
\label{fig:confusion}
\vspace{-2mm}

\end{figure}

\begin{figure}
	\centering
	\includegraphics[width=1\linewidth, height=0.17\textheight]{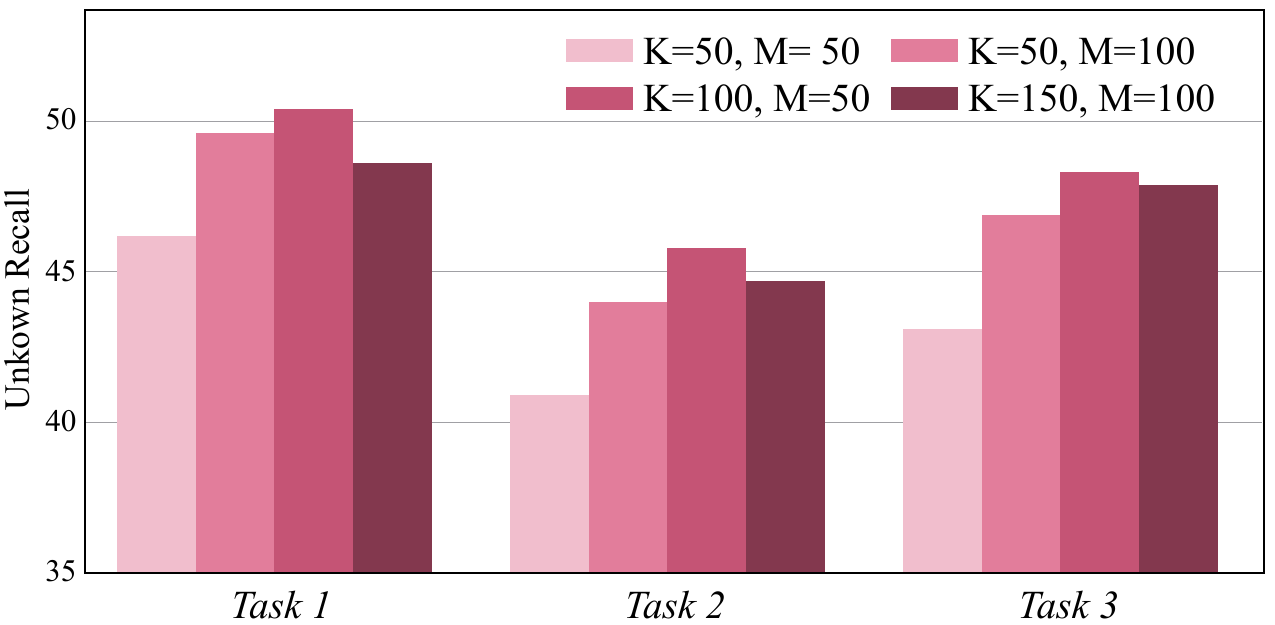}
\caption{Ablation study on the number of shared concepts on M-OWODB.
$K$ and $M$ denote the numbers of LLM-derived shared concepts and residual shared concepts, respectively.}
\label{fig:share_concept_num}
\vspace{-4mm}

\end{figure}

\subsection{Ablation Study}

% back up for 1.28 night
% \textbf{Effectiveness of Main Components.}
% Table~\ref{tab:ablation_mowodb} reports the contribution of each component of IPOW on M-OWODB.
% Adding the GMM-based RPN improves unknown recall by +4.1–4.7 points across Tasks~1–3 by enhancing proposal coverage for unseen objects, albeit at the cost of increased A-OSE due to less selective proposals.
% Discriminative concepts primarily benefit known-class recognition, improving known-class mAP by +0.4–1.6 points with marginal impact on U-Recall, indicating their role in strengthening class separation rather than unknown discovery.
% In contrast, introducing shared concepts yields substantial gains in U-Recall (+22.8–23.6 points), demonstrating that transferable semantic concepts are the key factors for generalizing knowledge to unseen categories.
% Further incorporating background concepts provides additional U-Recall improvements (+4.8–6.4 points) by explicitly modeling non-object semantics, which helps distinguish unknown objects from background clutter.
% Finally, CGR significantly reduces known--unknown confusion, decreasing WI by up to 61.9\% and A-OSE by up to 54.9\%, while preserving high U-Recall, highlighting its critical help in stabilizing open-world inference.

\textbf{Effectiveness of Main Components.}
Table~\ref{tab:ablation_mowodb} reports the contribution of each component of IPOW on M-OWODB.
Adding the GMM-based RPN improves unknown recall by 4.1–4.7 points across Tasks~1–3 by reducing the RPN’s bias toward known categories.
Discriminative concepts benefit known-class recognition, improving known-class mAP by 0.3–0.7 points, indicating their role in strengthening known class separation.
In contrast, introducing shared concepts yields substantial gains in U-Recall (18.3–22.1 points), demonstrating that shared semantic concepts are the key factors for generalizing knowledge to unseen categories.
Further incorporating background concepts provides additional U-Recall improvements (4.8–6.4 points) by explicitly modeling non-object semantics, which helps distinguish unknown objects from background clutter.
Finally, CGR significantly reduces known--unknown confusion, achieving a reduction of 19.7\%, 65.7\% and 60.2\% in WI and 41.7\%, 61.0\% and 57.5\% in A-OSE across 3 tasks, highlighting its critical role in resolving known--unknown confusion.

\begin{figure}
	\centering
	\includegraphics[width=1\linewidth]{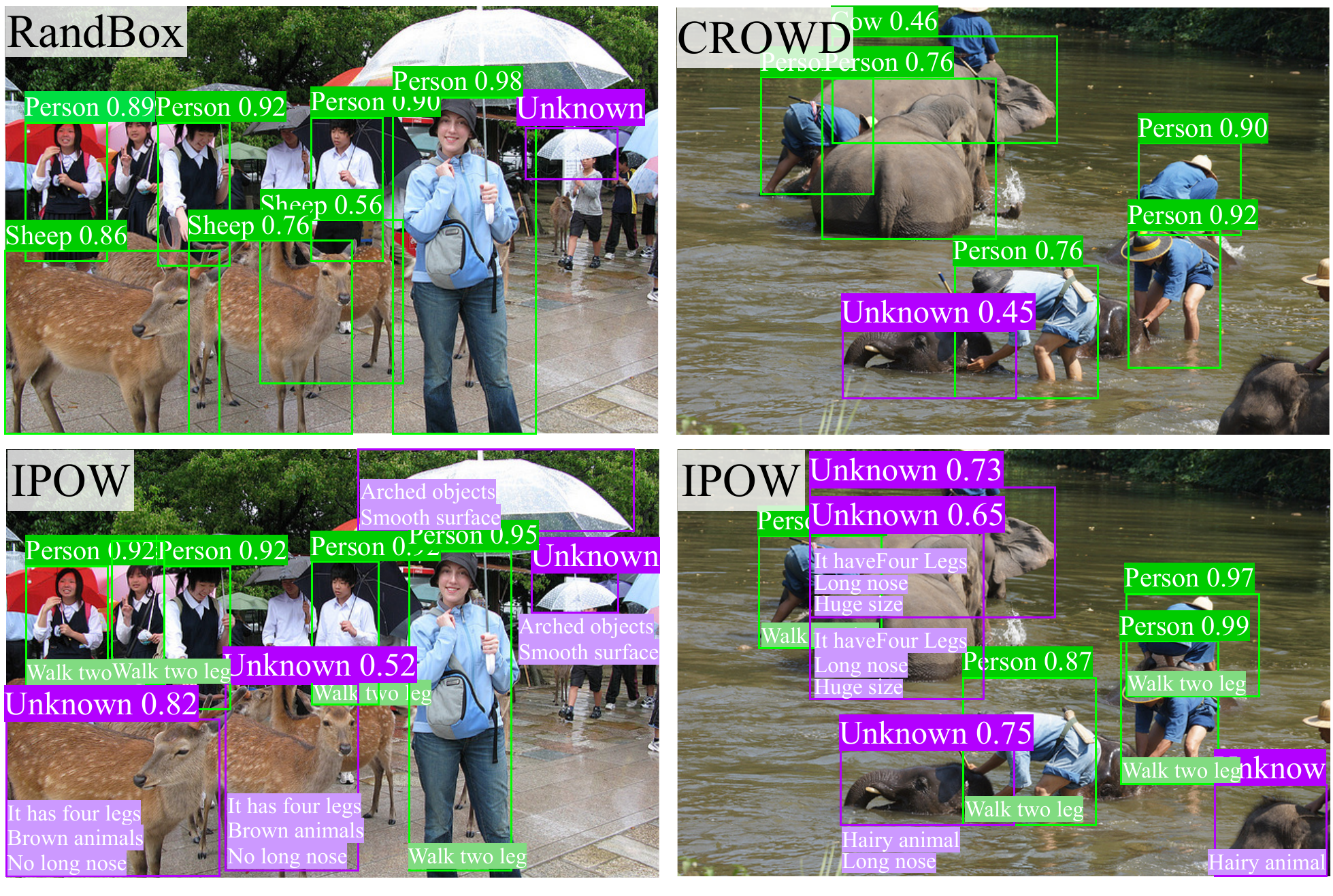}
\caption{Qualitative results contrasting IPOW with RandBox, and CROWD, demonstrating that IPOW offers interpretable concept-level reasoning, reduces known--unknown confusion, and enables effective generalization to unknown objects.}
\label{fig:visualization_v2}
\vspace{-4mm}
\end{figure}

\textbf{Statistical Analysis of Known-Unknown Confusion.}
As shown in Fig.~\ref{fig:confusion} (Left), those confused unknown objects (unknown falsely predicted as known categories) exhibit highly activated discriminative concepts, resembling known objects in the discriminative space. 
However, as illustrated in the shared concept group on the right, these confused unknown instances can be explicitly excluded. 

\textbf{Number of Shared Concepts.}
As shown in Fig.~\ref{fig:share_concept_num}, we vary the numbers of LLM-derived and residual shared concepts.
The best performance is achieved with $K=100$ and $M=50$, demonstrating that residual shared concepts effectively complement LLM-derived ones.
In contrast, increasing the numbers to $K=150$ and $M=100$ degrades recall, indicating that excessive shared concepts introduce redundancy and harm unknown detection.

\subsection{Qualitative Analysis}
As shown in Fig.~\ref{fig:visualization_v2}, IPOW provides a concept-level explanation for all known and unknown predictions with activated concepts.
For unknown objects, IPOW allows users to clearly localize objects of interest and identify relevant semantics.
This interpretability facilitates user annotation of unknown objects and their incorporation into subsequent tasks for incremental learning.
Compared with existing methods, IPOW shows higher unknown recall while effectively mitigating known–unknown confusion.

% 内部组件的消融
% 共享概念的数量 影响 unknown recall

% Concept-guided Rectification 修正的尺度

\section{Conclusions}

In this paper, we proposed IPOW, a concept-driven interpretable framework for open-world object detection.
Inspired by Concept Bottleneck Models, IPOW introduces a Concept Decomposition Model (CDM) that decomposes RoI features into discriminative, shared, and background concepts, enabling structured and interpretable reasoning over both known and unknown objects. Through this formulation, we reveal that known--unknown confusion arises when unknown objects fall into the discriminative space learned for known classes.
To address this issue, we further propose Concept-Guided Rectification (CGR), which leverages shared concept activations to effectively suppress such confusion in a principled and interpretable manner. Overall, IPOW demonstrates that concept-level decomposition provides a powerful solution for interpretable knowledge transfer in open-world detection, offering both improved reliability and transparent decision-making.
\section*{Impact Statement}

This paper presents work whose goal is to advance the field of machine learning. There are many potential societal consequences of our work, none of which we feel must be specifically highlighted here.


\bibliography{icml2026}
\bibliographystyle{icml2026}

%%%%%%%%%%%%%%%%%%%%%%%%%%%%%%%%%%%%%%%%%%%%%%%%%%%%%%%%%%%%%%%%%%%%%%%%%%%%%%%
%%%%%%%%%%%%%%%%%%%%%%%%%%%%%%%%%%%%%%%%%%%%%%%%%%%%%%%%%%%%%%%%%%%%%%%%%%%%%%%
% APPENDIX
%%%%%%%%%%%%%%%%%%%%%%%%%%%%%%%%%%%%%%%%%%%%%%%%%%%%%%%%%%%%%%%%%%%%%%%%%%%%%%%
%%%%%%%%%%%%%%%%%%%%%%%%%%%%%%%%%%%%%%%%%%%%%%%%%%%%%%%%%%%%%%%%%%%%%%%%%%%%%%%
\newpage
\appendix
\onecolumn
\begin{figure*}[t]
    \centering
    \includegraphics[width=\textwidth]{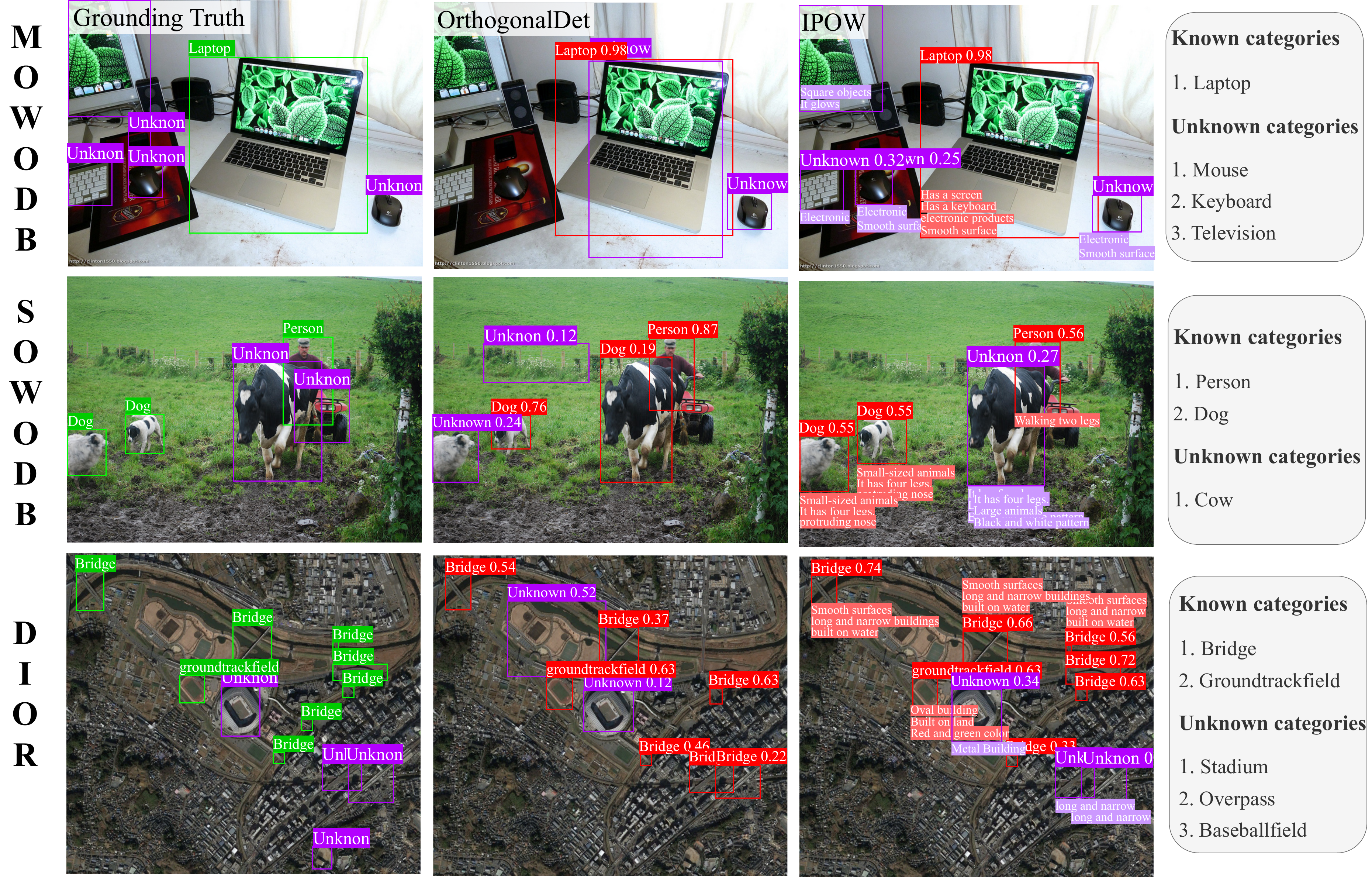}
\caption{Qualitative results comparing IPOW with OrthogonalDet on M-OWODB, S-OWODB, and the DIOR dataset, showing that IPOW provides interpretable concept-level reasoning and more effectively reduces known--unknown confusion across diverse scenarios.}
    \label{fig:visualization}
    \vspace{-4mm}
\end{figure*}

\begin{table*}[t]
\centering
\caption{Comparison with state-of-the-art methods in single-step incremental object detection settings on PASCAL VOC 2007.
Best results are shown in \textbf{bold}, while newly introduced classes in each task are shaded in gray.
Our method consistently outperforms existing approaches across all settings.}
\label{tab:iod_benchmark}
\setlength{\tabcolsep}{3pt}
\resizebox{\textwidth}{!}{%
\begin{tabular}{@{}lccccccccccccccccccccc@{}}
\toprule
% 10 + 10
{\textbf{10 + 10 setting}} & aero & cycle & bird & boat & bottle & bus & car & cat & chair & cow & table & dog & horse & bike & person & plant & sheep & sofa & train & tv & mAP \\ \midrule
ILOD \cite{Shmelkov_2017_ICCV} & 69.9 & 70.4 & 69.4 & 54.3 & 48 & 68.7 & 78.9 & 68.4 & 45.5 & 58.1 & \cellcolor{lightgray} 59.7 & \cellcolor{lightgray} 72.7 & \cellcolor{lightgray} 73.5 & \cellcolor{lightgray} 73.2 & \cellcolor{lightgray} 66.3 & \cellcolor{lightgray} 29.5 & \cellcolor{lightgray} 63.4 & \cellcolor{lightgray} 61.6 & \cellcolor{lightgray} 69.3 & \cellcolor{lightgray} 62.2 & 63.2 \\

Faster ILOD \cite{peng2020faster} & 72.8 & 75.7 & 71.2 & 60.5 & 61.7 & 70.4 & 83.3 & 76.6 & 53.1 & 72.3 & \cellcolor{lightgray} 36.7 & \cellcolor{lightgray} 70.9 & \cellcolor{lightgray} 66.8 & \cellcolor{lightgray} 67.6 & \cellcolor{lightgray} 66.1 & \cellcolor{lightgray} 24.7 & \cellcolor{lightgray} 63.1 & \cellcolor{lightgray} 48.1 & \cellcolor{lightgray} 57.1 & \cellcolor{lightgray} 43.6 & 62.1 \\ 
% \midrule
ORE ~\cite{joseph2021towards} & 63.5 & 70.9 & 58.9 & 42.9 & 34.1 & 76.2 & 80.7 & 76.3 & 34.1 & 66.1 & \cellcolor{lightgray} 56.1 & \cellcolor{lightgray} 70.4 & \cellcolor{lightgray} 80.2 & \cellcolor{lightgray} 72.3 & \cellcolor{lightgray} 81.8 & \cellcolor{lightgray} 42.7 & \cellcolor{lightgray} 71.6 & \cellcolor{lightgray} 68.1 & \cellcolor{lightgray} 77.0 & \cellcolor{lightgray} 67.7 & 64.5 \\
Meta-ILOD \cite{joseph2021incremental} & 76.0 & 74.6 & 67.5 & 55.9 & 57.6 & 75.1 & 85.4 &77.0 &43.7 &70.8 & \cellcolor{lightgray}  60.1 & \cellcolor{lightgray}  66.4 & \cellcolor{lightgray}  76.0 & \cellcolor{lightgray}  72.6 & \cellcolor{lightgray}  74.6 & \cellcolor{lightgray}  39.7 & \cellcolor{lightgray}  64.0 & \cellcolor{lightgray}  60.2 & \cellcolor{lightgray} 68.5 & \cellcolor{lightgray}  60.7 & 66.3 \\

ROSETTA \cite{yang2022continual} &74.2 & 76.2 & 64.9 & 54.4 &  57.4 & 76.1 & 84.4 & 68.8 & 52.4 &  67.0 & \cellcolor{lightgray}  62.9 & \cellcolor{lightgray}   63.3 & \cellcolor{lightgray}   79.8 &\cellcolor{lightgray}  72.8 & \cellcolor{lightgray}  78.1 & \cellcolor{lightgray}  40.1 &\cellcolor{lightgray}  62.3 & \cellcolor{lightgray}  61.2 & \cellcolor{lightgray}  72.4 & \cellcolor{lightgray}  66.8 & 66.8 \\

OW-DETR\cite{gupta2022ow} &61.8 & 69.1 & 67.8 & 45.8 & 47.3 & 78.3 &78.4 &78.6 &36.2 & 71.5 & \cellcolor{lightgray}  57.5 & \cellcolor{lightgray}  75.3 & \cellcolor{lightgray}  76.2 &\cellcolor{lightgray}  77.4 &\cellcolor{lightgray}  79.5 &\cellcolor{lightgray}  40.1 &\cellcolor{lightgray}  66.8 &\cellcolor{lightgray}  66.3 &\cellcolor{lightgray}  75.6 &\cellcolor{lightgray}  64.1 & 65.7 \\

PROB \cite{zohar2023prob} &70.4 & 75.4 & 67.3 & 48.1 & 55.9 & 73.5 & 78.5 & 75.4 & 42.8 & 72.2 & \cellcolor{lightgray}  64.2 &\cellcolor{lightgray}  73.8 &\cellcolor{lightgray}  76.0 & \cellcolor{lightgray}  74.8 &\cellcolor{lightgray}  75.3 &\cellcolor{lightgray}  40.2 &\cellcolor{lightgray} 66.2 &\cellcolor{lightgray}  73.3 &\cellcolor{lightgray}  64.4 &\cellcolor{lightgray}  64.0 & 66.5 \\

CAT \cite{ma2023cat} & 76.5 & 75.7 & 67.0 & 51.0 & 62.4 & 73.2 & 82.3 & 83.7 & 42.7 & 64.4 & \cellcolor{lightgray}  56.8 &\cellcolor{lightgray}  74.1 &\cellcolor{lightgray}  75.8 &\cellcolor{lightgray}  79.2 &\cellcolor{lightgray}  78.1 &\cellcolor{lightgray}  39.9 &\cellcolor{lightgray}  65.1 &\cellcolor{lightgray}  59.6 &\cellcolor{lightgray}  78.4 &\cellcolor{lightgray}  67.4 & 67.7 \\

OrthogonalDet \cite{sun2024exploring}\footnotemark[1] &82.9 &80.1 &75.8 &64.3 &60.6 &81.5 & 87.9 & 54.9 & 48 &82.1 &\cellcolor{lightgray}  57.7 & \cellcolor{lightgray} 63.5 &\cellcolor{lightgray}  80.5 &\cellcolor{lightgray} 77.6 &\cellcolor{lightgray} 78.2 &\cellcolor{lightgray} 38.9 &\cellcolor{lightgray} 69.8 &\cellcolor{lightgray} 62.8 &\cellcolor{lightgray} 76.9 &\cellcolor{lightgray} 64.2 &69.41 \\

CROWD \cite{majee2025looking} & 84.1 & 84.5 & 73.9 & 60.0 &	65.1 & 80.1 & 89.3 & 82.7 &	53.3 & 77.4 & \cellcolor{lightgray}63.4 & \cellcolor{lightgray}78.5 &	\cellcolor{lightgray}80.9 & \cellcolor{lightgray}83.4 & \cellcolor{lightgray}83.9 & \cellcolor{lightgray}46.5 & \cellcolor{lightgray}72.6 & \cellcolor{lightgray}60.9 & \cellcolor{lightgray}77.9 & \cellcolor{lightgray}71.5 &	73.5 \\

\midrule

\textbf{IPOW (ours)} & 79.8 & 83.7 & 78.0 & 63.1 &	65.1 & 81.3 & 87.0 & 84.8 &	62.0 & 71.3 & \cellcolor{lightgray}64.9 & \cellcolor{lightgray}82.5 &	\cellcolor{lightgray}79.0 & \cellcolor{lightgray}79.1 & \cellcolor{lightgray}83.8 & \cellcolor{lightgray}48.6 & \cellcolor{lightgray}66.0 & \cellcolor{lightgray}67.8 & \cellcolor{lightgray}74.9 & \cellcolor{lightgray}74.2 &	\textbf{73.8} \\

\midrule\midrule
% 15 + 5
{\textbf{15 + 5 setting}} & aero & cycle & bird & boat & bottle & bus & car & cat & chair & cow & table & dog & horse & bike & person & plant & sheep & sofa & train & tv & mAP \\ \midrule
ILOD \cite{Shmelkov_2017_ICCV} & 70.5 & 79.2 & 68.8 & 59.1 & 53.2 & 75.4  & 79.4 & 78.8 & 46.6 & 59.4 & 59.0 & 75.8 & 71.8 & 78.6 & 69.6 & \cellcolor{lightgray}  33.7 & \cellcolor{lightgray}  61.5 & \cellcolor{lightgray}  63.1 &\cellcolor{lightgray}  71.7 &\cellcolor{lightgray}  62.2 & 65.8 \\

Faster ILOD \cite{peng2020faster} & 66.5 & 78.1 & 71.8 & 54.6 & 61.4 & 68.4 & 82.6 & 82.7 & 52.1 & 74.3 & 63.1 & 78.6 & 80.5 & 78.4 & 80.4 & \cellcolor{lightgray} 36.7 &\cellcolor{lightgray}  61.7 &\cellcolor{lightgray}  59.3 &\cellcolor{lightgray}  67.9 &\cellcolor{lightgray}  59.1 & 67.9 \\
% \midrule
ORE ~\cite{joseph2021towards} & 75.4 & 81.0 & 67.1 & 51.9 & 55.7 & 77.2 & 85.6 & 81.7 & 46.1 & 76.2 & 55.4 & 76.7 & 86.2 & 78.5 & 82.1 &\cellcolor{lightgray}  32.8 & \cellcolor{lightgray}  63.6 &\cellcolor{lightgray}  54.7  &\cellcolor{lightgray} 77.7 &\cellcolor{lightgray}  64.6 & 68.5 \\
Meta-ILOD \cite{joseph2021incremental} & 78.4 & 79.7 & 66.9 & 54.8 & 56.2 & 77.7 & 84.6 & 79.1 & 47.7 & 75.0 & 61.8 & 74.7 & 81.6 & 77.5 & 80.2 &\cellcolor{lightgray}  37.8 &\cellcolor{lightgray}  58.0 &\cellcolor{lightgray}  54.6 &\cellcolor{lightgray}  73.0 &\cellcolor{lightgray}  56.1 & 67.8 \\
ROSETTA \cite{yang2022continual} & 76.5 & 77.5 & 65.1 & 56.0 & 60.0 & 78.3 & 85.5 & 78.7 & 49.5 & 68.2 & 67.4 & 71.2 & 83.9 & 75.7 & 82.0 &\cellcolor{lightgray}   43.0 &\cellcolor{lightgray}  60.6 &\cellcolor{lightgray}  64.1 &\cellcolor{lightgray}  72.8 &\cellcolor{lightgray}  67.4 & 69.2 \\
OW-DETR \cite{gupta2022ow}& 77.1 & 76.5 & 69.2 & 51.3 & 61.3 & 79.8 & 84.2 & 81.0 & 49.7 & 79.6 & 58.1 & 79.0 & 83.1 & 67.8 & 85.4 & \cellcolor{lightgray} 33.2 & \cellcolor{lightgray} 65.1 & \cellcolor{lightgray} 62.0 & \cellcolor{lightgray} 73.9 & \cellcolor{lightgray} 65.0 & 69.4 \\ 
PROB \cite{zohar2023prob} & 77.9 & 77.0 & 77.5 & 56.7 & 63.9 & 75.0 & 85.5 & 82.3 & 50.0 & 78.5 & 63.1 & 75.8 & 80.0 & 78.3 & 77.2 &\cellcolor{lightgray}  38.4 &\cellcolor{lightgray}  69.8 &\cellcolor{lightgray}  57.1 &\cellcolor{lightgray}  73.7 &\cellcolor{lightgray}  64.9 & 70.1 \\
CAT \cite{ma2023cat} &75.3  &81.0 & 84.4 & 64.5 & 56.6  &74.4 & 84.1 & 86.6 & 53.0 & 70.1 & 72.4 & 83.4 & 85.5 & 81.6 & 81.0 &\cellcolor{lightgray} 32.0 &\cellcolor{lightgray}  58.6 &\cellcolor{lightgray}  60.7 &\cellcolor{lightgray}  81.6 &\cellcolor{lightgray}  63.5 & 72.2 \\
OrthogonalDet \cite{sun2024exploring}\footnotemark[1] & 81.8 & 79.3 & 71.0 & 71.0 & 58.8 & 62.1 & 82.6 & 89.7 & 79.8 & 47.0 & 80.5 & 61.1 & 79.9 & 80.2 & 81.6 & \cellcolor{lightgray}44.2 &\cellcolor{lightgray}65.5 &\cellcolor{lightgray}71.5 &\cellcolor{lightgray}75.6 &\cellcolor{lightgray} 74.2 & 72.6 \\
CROWD\cite{majee2025looking} &  82.8 & 80.6 & 72.5 &59.6 &61.3 &83.1 &89.3 &83 &49.2 &86.1 &62.2 &83.7 &86 &80.3 &82.8 & \cellcolor{lightgray} 46.1 &\cellcolor{lightgray} 80 &\cellcolor{lightgray} 63.7 &\cellcolor{lightgray} 79.5 &\cellcolor{lightgray} 75.6 & 74.4 \\ 

\midrule

\textbf{IPOW (ours)} &  77.9 & 79.5 & 77.0 &65.7 &63.0 &82.2 &86.2 &87.7 &56.6 &82.8 &68.9 &86.7 &85.9 &79.6 &84.8 & \cellcolor{lightgray} 44.6 &\cellcolor{lightgray} 71.7 &\cellcolor{lightgray} 68.6 &\cellcolor{lightgray} 74.6 &\cellcolor{lightgray} 75.2 & \textbf{75.0} \\ 
\midrule\midrule

% 19 + 1
{\textbf{19 + 1 setting}} & aero & cycle & bird & boat & bottle & bus & car & cat & chair & cow & table & dog & horse & bike & person & plant & sheep & sofa & train & tv & mAP \\ \midrule
ILOD \cite{Shmelkov_2017_ICCV} & 69.4 & 79.3 & 69.5 & 57.4 & 45.4 & 78.4 & 79.1 & 80.5 & 45.7 & 76.3 & 64.8 & 77.2 & 80.8 & 77.5 & 70.1 & 42.3 & 67.5 & 64.4 & 76.7 & \cellcolor{lightgray} 62.7 & 68.2 \\

Faster ILOD \cite{peng2020faster} & 64.2 & 74.7 & 73.2 & 55.5 & 53.7 & 70.8 & 82.9 & 82.6 & 51.6 & 79.7 & 58.7 & 78.8 & 81.8 & 75.3 & 77.4 & 43.1 & 73.8 & 61.7 & 69.8 & \cellcolor{lightgray} 61.1 & 68.5 \\

ORE ~\cite{joseph2021towards} & 67.3 & 76.8 & 60 & 48.4 & 58.8 & 81.1 & 86.5 & 75.8 & 41.5 & 79.6 & 54.6 & 72.8 & 85.9 & 81.7 & 82.4 & 44.8 & 75.8 & 68.2 & 75.7 & \cellcolor{lightgray} 60.1 & 68.8 \\ 

Meta-ILOD \cite{joseph2021incremental} &78.2 & 77.5 & 69.4 & 55.0 & 56.0 & 78.4 & 84.2 & 79.2 & 46.6 & 79.0 & 63.2 & 78.5 & 82.7 & 79.1 & 79.9 & 44.1 & 73.2 & 66.3 & 76.4 &\cellcolor{lightgray}  57.6 & 70.2 \\

ROSETTA \cite{yang2022continual} & 75.3 & 77.9 & 65.3 & 56.2 & 55.3 & 79.6 &84.6 &72.9 &49.2 &73.7 &68.3 &71.0 &78.9 &77.7 &80.7 &44.0& 69.6 &68.5& 76.1& \cellcolor{lightgray} 68.3 &69.6 \\

OW-DETR \cite{gupta2022ow} & 70.5 & 77.2 & 73.8 & 54.0 & 55.6 & 79.0 & 80.8 & 80.6 & 43.2 & 80.4 & 53.5 & 77.5 & 89.5 & 82.0 & 74.7 & 43.3 & 71.9 & 66.6 & 79.4 & \cellcolor{lightgray} 62.0 & 70.2 \\

PROB \cite{zohar2023prob} & 80.3 &78.9 &77.6 &59.7 &63.7 &75.2 &86.0 &83.9 &53.7 &82.8 &66.5 &82.7 &80.6 &83.8 &77.9 &48.9 &74.5 &69.9 &77.6  &\cellcolor{lightgray}  48.5 &   72.6 \\

CAT \cite{ma2023cat} &86.0 &85.8 &78.8 &65.3 &61.3 &71.4 &84.8 &84.8 &52.9 &78.4 &71.6 &82.7 &83.8 &81.2 &80.7 &43.7 &75.9 &58.5 &85.2 &\cellcolor{lightgray} 61.1 & 73.8 \\
OrthogonalDet \cite{sun2024exploring}\footnotemark[1] & 81.8 & 82.6 & 77.0 & 56.3 & 66.0 & 74.4 & 88.5 & 78.7 & 51.2 & 84.3 & 63.1 & 84.4 & 81.3 & 78.8 & 80.9 & 46.8 & 77.9 & 68.6 & 74.1 &\cellcolor{lightgray} 74.5 & 73.6 \\
CROWD \cite{majee2025looking}&  81.7 &	80.3 & 77.4 & 57.2 & 66.8 & 80.7 &	87.1 & 67.9 & 49.4 & 87.3 & 65.6 & 84.2 & 85.4 & 79.9 & 81.6 & 48.6 & 77.0 & 69.0 & 82.2 & \cellcolor{lightgray}75.3 & 74.2 \\ 
\midrule
\textbf{IPOW (ours)} &  81.1 &	78.7 & 78.2 & 56.3 & 63.5 & 77.2 & 86.2 & 86.8 &  59.5 & 79.8 & 64.7 & 86.3 & 83.0 & 78.8 & 83.2 & 49.3 & 74.4 & 71.6 & 81.0 & \cellcolor{lightgray}71.0 & \textbf{74.5} \\ 
\bottomrule
\end{tabular}%
}

\end{table*}

\section{Visualization}

As shown in Fig.~\ref{fig:visualization}, we qualitatively compare IPOW with OrthogonalDet on M-OWODB, S-OWODB, and the remote sensing DIOR dataset.
The results highlight the interpretability of IPOW, where both known and unknown predictions are explained through activated semantic concepts.

IPOW effectively mitigates known--unknown confusion.
For instance, in S-OWODB, visually similar objects such as \emph{cow} are misclassified as the known class \emph{dog} by OrthogonalDet, whereas IPOW excludes such instances from known categories based on incomplete shared concept activations, correctly identifying them as unknown.
This provides a clear concept-level explanation of why these instances should not be classified as known.

Furthermore, results on DIOR show that IPOW remains effective in remote sensing scenarios with distributions markedly different from natural images, producing reliable unknown predictions with interpretable concept responses.

\section{Experiment setup details}
\label{appsec:experiment_details}

\paragraph{Datasets}
We follow the common evaluation protocol~\cite{sun2024exploring,wang2023random} and evaluate all methods on both open-world and incremental object detection benchmarks.
For open-world object detection, we consider the superclass-mixed benchmark M-OWODB~\cite{joseph2021towards} and the superclass-separated benchmark S-OWODB~\cite{gupta2022ow}.
M-OWODB combines PASCAL VOC and COCO, where all VOC categories are treated as known classes in Task~1, and the remaining COCO categories are introduced incrementally as unknown classes.
In contrast, S-OWODB uses only the COCO dataset and strictly separates superclasses across tasks, resulting in more isolated semantic partitions and a smaller amount of training data compared to M-OWODB. For incremental object detection, we adopt the class splits of VOC 2007 proposed in~\cite{Shmelkov_2017_ICCV}, which include multiple two-stage incremental settings (10+10, 15+5, and 19+1).

To further investigate whether IPOW remains effective under scenarios with significantly different data distributions from common daily scenes such as COCO and PASCAL VOC, we conduct open-world experiments on the popular remote sensing dataset DOIR~\cite{li2020object}.
DOIR is a remote sensing object detection benchmark consisting of 20 object categories with diverse aerial viewpoints.
We partition the dataset into two tasks according to the alphabetical order of category names, with 10 classes in each task.
During training and evaluation of the first task, the classes belonging to the second task are treated as unknown categories.

\paragraph{Implementation Details}  
Our method is built upon Faster R-CNN~\cite{ren2016faster} and employs a ResNet-50~\cite{he2016deep} backbone pretrained on ImageNet, without building upon any existing OWOD methods.
The model is trained using the SGD optimizer with a learning rate of 0.02 and a batch size of 16.
% All experiments are conducted on 4 NVIDIA RTX 3090 GPUs.
For the shared concept space, the numbers of LLM-derived and residual shared concepts are set to $K=100$ and $M=50$, respectively.
In the concept-guided rectification module, the rectification strength parameter $\eta$ is set to 0.8.
Note that in our method, the LLM is used only for semantic modeling when new tasks are introduced and does not participate in training or inference, incurring no additional overhead.

\section{GMM RPN}
\label{appsec:gmm_rpn}

In Faster R-CNN, the RPN generates candidates but suffers from a strong bias toward known categories, as unannotated unknown objects are often suppressed as background. While sampling proposals from random Gaussian noise \cite{wang2023random} can mitigate this bias, it sacrifices the RPN's refinement capability and necessitates multiple forward passes, significantly increasing inference overhead. 

We propose an efficient GMM-based strategy based on the observation that objects, regardless of category, exhibit consistent spatial and scale priors: they are more frequently located toward the center of the image and occupy medium-sized bounding boxes. This pattern naturally aligns with a Gaussian distribution. Thus, we employ a Gaussian Mixture Model (GMM) to fit the distribution of known ground-truth boxes, capturing universal objectness priors that generalize to the open world without additional cost. Formally, we represent each bounding box as $\mathbf{b} = [x, y, w, h]$ and model the spatial-scale distribution of known categories using a Gaussian Mixture Model (GMM):
\begin{equation}
    P(\mathbf{b}) = \sum_{k=1}^{K} \pi_{k} \mathcal{N}(\mathbf{b} | \boldsymbol{\mu}_{k}, \boldsymbol{\Sigma}_{k}),
\end{equation}
where $\pi_{k}$, $\boldsymbol{\mu}_{k}$, and $\boldsymbol{\Sigma}_{k}$ denote the mixing weights, means, and covariances of the $k$-th Gaussian component, respectively. 
By modeling intrinsic geometric priors, the GMM captures universal object distributions that mitigate known-category bias. We retain the original learnable RPN for precise known-class detection and use GMM-sampled proposals as a zero-cost complement, providing spatially informed candidates for both known and unknown objects without additional inference overhead.

% IOD

%

\section{Incremental Object Detection}

Incremental Object Detection (IOD), as a subtask of OWOD, 
aims to incrementally acquire new object categories over time while retaining the ability to detect previously learned ones.
In this way, OWOD is able to discover categories of interest from unknown objects and learn them progressively, enabling the detector to continuously evolve in real-world scenarios.

Previous works following OW-DETR~\cite{gupta2022ow} employ exemplar replay-based fine-tuning to alleviate catastrophic forgetting.
However, as noted in~\cite{zhang2025yolo}, catastrophic forgetting in incremental object detection stems from a more fundamental issue, namely \emph{foreground–background confusion}, where unannotated objects from previously learned classes are misclassified as background during training.
Exemplar replay alone cannot effectively address this problem and also incurs additional memory and computational overhead.
Instead, following~\cite{zhang2025yolo}, we adopt a pseudo-labeling strategy that leverages models from previous tasks to generate pseudo labels for previously learned classes, thereby mitigating foreground–background confusion during incremental learning.

\section{IOD Benchmark Results}

Explicitly identifying unknown objects transforms the abrupt introduction of new categories into a progressive learning process from previously discovered unknown instances, enabling better knowledge preservation during adaptation to new tasks. In IPOW, strong unknown detection capability is achieved, while known categories are detected through discriminative concepts organized into an Equiangular Tight Frame (ETF) structure.
During incremental learning, new categories are separated by additional discriminative concepts without altering the existing discriminative space, allowing most previously acquired knowledge to be preserved.

As shown in Table~\ref{tab:iod_benchmark}, IPOW consistently outperforms existing methods across all single-step incremental settings.
Specifically, IPOW surpasses CROWD by 0.3, 0.6, and 0.3 mAP under the 10+10, 15+5, and 19+1 settings, achieving mAP scores of 73.8, 75.0, and 74.5, respectively.
These results demonstrate that the proposed concept-based incremental knowledge transfer strategy effectively alleviates catastrophic forgetting in incremental object detection.
\section{LLM-Based Semantic Concept Construction}
% 判别概念语义生成, 给定一组已知类别 生成所有的 2 个类别组合, 利用下述 prompt
% Identify ONE the most discriminative visual attribute 对 betweent  ,例如：人有两条腿  猫有四条腿.
% 这样就生成类别组合间 最具判别性的属性
% 当新任务引入一组新的类别, 我们要计算 新类别之前的 判别特征与新类别与先前类别的判别特征 方法同上

\textbf{Discriminative Concepts Construction.}
As summarized in Algorithm~\ref{alg:disc_concepts}, we construct discriminative concepts via LLM-assisted semantic generation.
Specifically, for a set of known classes, we enumerate all unordered class pairs and query an LLM with a fixed prompt to obtain one most discriminative visual attribute for each pair.
Specifically, we query the LLM with the discriminative prompt $\mathcal{P}_{dis}(c_i,c_j)$:
\emph{``Between the following two object classes, identify ONE most discriminative visual attribute that strictly differentiates them.''}
When new classes are introduced, the discriminative concept set is incrementally extended by applying the same pairwise procedure between newly introduced and previously learned classes, while keeping existing discriminative concepts unchanged.

\begin{algorithm}[t]
\caption{Discriminative Concepts Construction}
\label{alg:disc_concepts}
\small
\begin{algorithmic}[1]
\STATE \textbf{Input:} known classes at task $t$: $\mathcal{K}_t$; previous known classes $\mathcal{K}_{t-1}$ (empty if $t{=}1$); LLM prompt template $\mathcal{P}_{dis}(\cdot,\cdot)$
\STATE \textbf{Output:} discriminative concept set $\mathcal{C}^{u}_t$; pair-to-attribute map $\mathcal{M}_t$

\STATE $\mathcal{C}^{u}_t \leftarrow \mathcal{C}^{u}_{t-1}$, \ $\mathcal{M}_t \leftarrow \mathcal{M}_{t-1}$ \ \ \textit{(if $t{=}1$, initialize as empty)}

\STATE $\mathcal{P}_{\mathrm{pair}} \leftarrow \{(c_i,c_j)\mid c_i,c_j\in\mathcal{K}_t,\ i<j\}$ \ \textit{(all pairs)}

\IF{$t >1$}
    \STATE $\mathcal{P}_{\mathrm{pair}} \leftarrow \mathcal{P}_{\mathrm{pair}} \cup \{(c_n,c_o)\mid c_n\in\mathcal{K}_t,\ c_o\in\mathcal{K}_{t-1}\}$ \ \textit{(new vs.\ old)}
\ENDIF

\FOR{each $(c_i,c_j)\in\mathcal{P}_{\mathrm{pair}}$}
    \STATE $a_{ij} \leftarrow \mathrm{LLM}\!\left(\mathcal{P}(c_i,c_j)\right)$
    \STATE $\mathcal{M}_t[(c_i,c_j)] \leftarrow a_{ij}$
    \STATE $\mathcal{C}^{u}_t \leftarrow \mathcal{C}^{u}_t \cup \{a_{ij}\}$
\ENDFOR

\STATE \textbf{return} $\mathcal{C}^{u}_t$, $\mathcal{M}_t$
\end{algorithmic}
\end{algorithm}

\textbf{Shared Concepts Construction.}
% LLM-derived  共享概念的生成, 我们通过两步 反演 query 来进行，具体来说 给定一组已知类别 我们生成 两两 的对，通过 通过prompt: 'List at least {num} shared visual attributes that BOTH categories have.' 先生成 部分共享的属性，因为该属性可能不仅仅属于该类别对，我们对于每个属性进行反演推理，利用LLM在当前类别中总结具有该属性的类别. promtp : Which of these classes possess the attribute '{attribute}'?
% 当新任务引入 我们仅对新任务重新生成 共享属性 添加到原有属性集中

As summarized in Algorithm~\ref{alg:shared_concepts}, we construct LLM-derived shared concepts using a two-step reverse-query process.
Given a set of known classes, we first enumerate all unordered class pairs and query the LLM with the pairwise prompt
$\mathcal{P}_{\mathrm{pair}}(c_i,c_j)$:
\emph{``List at least $n$ shared visual attributes that BOTH categories have.''}
to generate shared visual attributes for each class pair.
Since a shared attribute may apply to more than the queried pair, we further perform attribute inversion: for each generated attribute $a$, we query the LLM with the inversion prompt
$\mathcal{P}_{\mathrm{inv}}(a)$:
\emph{``Which of these classes possess the attribute `$\{a\}$'?''}
to identify the set of classes in the current known category set that exhibit this attribute, yielding an attribute-to-classes mapping.
When a new task is introduced, newly generated shared attributes and their corresponding mappings are merged into the existing shared concept set.

\begin{algorithm}[t]
\caption{Shared Concepts Construction (LLM-Derived)}
\label{alg:shared_concepts}
\small
\begin{algorithmic}[1]
\STATE \textbf{Input:} known classes at task $t$: $\mathcal{K}_t$; prompts $\mathcal{P}_{\mathrm{pair}}(\cdot,\cdot)$ and $\mathcal{P}_{\mathrm{inv}}(\cdot)$
\STATE \textbf{Output:} shared concept set $\mathcal{C}^{v}_t$; attribute-to-classes map $\Phi_t$

\STATE $\mathcal{C}^{v}_t \leftarrow \mathcal{C}^{v}_{t-1}$, \ $\Phi_t \leftarrow \Phi_{t-1}$ \ \ \textit{(if $t{=}1$, initialize as empty)}
\STATE $\mathcal{A}_{\mathrm{new}} \leftarrow \emptyset$

\STATE $\mathcal{P}_{\mathrm{pair}} \leftarrow \{(c_i,c_j)\mid c_i,c_j\in\mathcal{K}_t,\ i<j\}$ \ \textit{(class pairs)}

\FOR{each $(c_i,c_j)\in\mathcal{P}_{\mathrm{pair}}$}
    \STATE $\mathcal{A}_{ij} \leftarrow \mathrm{LLM}\!\left(\mathcal{P}_{\mathrm{pair}}(c_i,c_j)\right)$ \ \textit{(shared attributes)}
    \STATE $\mathcal{A}_{\mathrm{new}} \leftarrow \mathcal{A}_{\mathrm{new}} \cup \mathcal{A}_{ij}$
\ENDFOR

\FOR{each attribute $a \in \mathcal{A}_{\mathrm{new}}$}
    \STATE $\mathcal{S}_a \leftarrow \mathrm{LLM}\!\left(\mathcal{P}_{\mathrm{inv}}(a)\right)$ \ \textit{(attribute inversion)}
    \STATE $\Phi_t[a] \leftarrow \mathcal{S}_a \cap \mathcal{K}_t$
\ENDFOR

\STATE $\mathcal{C}^{v}_t \leftarrow \mathcal{C}^{v}_t \cup \mathcal{A}_{\mathrm{new}}$
\STATE \textbf{return} $\mathcal{C}^{v}_t$, $\Phi_t$
\end{algorithmic}
\end{algorithm}

\end{document}